\documentclass{article}

\usepackage[utf8]{inputenc}
\usepackage{amsmath, amsthm, amssymb, amsfonts, dsfont, mathrsfs, dirtytalk, tikz-cd, adjustbox, url, nicefrac, booktabs,array, graphicx,todonotes,longtable, tabu,bm,cite, todonotes,comment,multirow,tabularx,epsfig,parskip,setspace}
\usepackage{caption}
\usepackage[T1]{fontenc}    
\usepackage[margin=1in]{geometry}
\usepackage[english]{babel}
\usepackage{hyperref}
\usepackage[capitalise]{cleveref}

\newcommand{\e}{\varepsilon}

\newcommand{\E}{\mathbb E}

\newcommand{\G}{\operatorname G}
\newcommand{\F}{\operatorname F}

\newcommand{\dkl}{\operatorname{D_{K L}}}

\theoremstyle{plain}
\newtheorem{theorem}{Theorem}[section]

\theoremstyle{definition}
\newtheorem{definition}{Definition}

\newtheorem{remark}[theorem]{Remark}

\theoremstyle{remark}

\newcounter{daggerfootnote}

\let\OLDthebibliography\thebibliography
\renewcommand\thebibliography[1]{
  \OLDthebibliography{#1}
  \setlength{\parskip}{0pt}
  \setlength{\itemsep}{0pt plus 0.3ex}
}

\title{Possible Principles for Aligned Structure Learning Agents}
\author{Lancelot Da Costa$^{1}$, Tomáš Gavenčiak$^2$, David Hyland$^{3}$, Mandana Samiei$^{4,5}$,\\ Cristian Dragos-Manta$^{4,6}$, Candice Pattisapu, 
Adeel Razi$^{7,8,9,10}$, Karl Friston$^{1,7}$}

\date{\small\it
    $^1$VERSES AI Research Lab, Los Angeles,
    USA\\
    $^2$Charles University, Prague, Czech Republic\\
    $^3$Department of Computer Science, University of Oxford, UK\\
    $^4$Mila, Quebec AI Institute, Montreal, Canada\\
    $^5$Department of Computer Science, McGill University, Montreal, Canada\\
    $^6$University of Montreal, Montreal, Canada\\
    $^7$Queen Sq, Institute of Neurology, University College London, London, 
    UK\\
    $^8$School of Psychological Sciences, Monash University, Clayton, Australia\\
    $^9$Monash Biomedical Imaging, Monash University, Clayton, Australia\\
    $^{10}$CIFAR Azrieli Global Scholars Program, Toronto, Canada\\
}
\begin{document}
\maketitle

\begin{abstract}
\normalsize
This paper offers a roadmap for the development of scalable aligned artificial intelligence (AI) from first principle descriptions of natural intelligence.
In brief, a possible path toward scalable aligned AI rests upon enabling artificial agents to learn a good model of the world that includes a good model of our preferences.
For this, the main objective is creating agents that learn to represent the world and other agents' world models; a problem that falls under structure learning (a.k.a. causal representation learning or model discovery).
We expose the structure learning and alignment problems with this goal in mind, as well as principles to guide us forward, synthesizing various ideas across mathematics, statistics, and cognitive science.
1)~We discuss the essential role of core knowledge, information geometry and model reduction in structure learning, and suggest core structural modules to learn a wide range of naturalistic worlds.
2)~We outline a way toward aligned agents through structure learning and theory of mind. As an illustrative example, we mathematically sketch Asimov's Laws of Robotics, which prescribe agents to act cautiously to minimize the ill-being of other agents. We supplement this example by proposing refined approaches to alignment.
These observations may guide the development of artificial intelligence in helping to scale existing---or design new---aligned structure learning systems.
\end{abstract}


\textbf{Keywords:} Agent, world model, generative, model discovery, causal representation learning, Bayesian inference.

\tableofcontents

\section{Introduction}

This paper examines the challenge of developing scalable aligned AI agents following biomimetic principles. We consider the research questions to be addressed, along with guiding principles; providing a broad perspective that synthesizes ideas across mathematics, physics, statistics, and cognitive science.

\paragraph{A first principles approach to intelligence:} We aim to be inclusive about---and relevant to---all naturalistic approaches to artificial intelligence. We approach this with a `first principles' approach to modeling intelligence known as the active inference framework \cite{dacostaActiveInferenceDiscrete2020,parrActiveInferenceFree2022,buckleyFreeEnergyPrinciple2017}. Active inference is not divorced from other naturalistic approaches to modeling intelligence, but rather aims to accommodate them within a broad framework derived from statistical physics. 
This follows a long lineage of ideas, perhaps originating from Helmholtz's motion of perception as unconscious inference \cite{helmholtzHelmholtzsTreatisePhysiological1962}, which was reincarnated in the neurosciences at the turn of this century as predictive coding \cite{raoPredictiveCodingVisual1999}, and generalized as the Bayesian brain hypothesis \cite{knillBayesianBrainRole2004}. 
Active inference was proposed shortly afterward, in the mid 2000s, extending these Bayesian accounts by postulating that action optimizes the same objective as perception and learning \cite{fristonTheoryCorticalResponses2005,fristonFreeenergyBrain2007}. This account was suggested to be a potentially unifying brain theory, in the sense of accommodating a wide range of previously existing and partially non-overlapping brain theories as special cases \cite{fristonFreeenergyPrincipleUnified2010}.
In light of the descriptive power of these ideas, researchers have sought to justify this account in terms of statistical physics, with increasing mathematical rigor and sophistication.
These efforts have birthed a nascent field of non-equilibrium physics, known as `Bayesian mechanics', that bridges stochastic descriptions of particles with inferential ones. This has been used to derive the active inference framework as we present it here, which provides a description of sentient behaviour \cite{fristonFreeEnergyPrinciple2023a}.
The active inference framework can be used to model a remarkable range of phenomena in cognitive science, ranging from human choice behavior \cite{parrComputationalNeurologyMovement2021a} to psychopathology \cite{adamsComputationalAnatomyPsychosis2013}, to many known features of the brain’s
anatomy and physiology \cite{fristonActiveInferenceProcess2017,fristonGraphicalBrainBelief2017}, including the activity of neural populations \cite{isomuraCanonicalNeuralNetworks2022,isomuraExperimentalValidationFreeenergy2023} (see \cite{dacostaActiveInferenceDiscrete2020,smithStepbystepTutorialActive2022,parrActiveInferenceFree2022,buckleyFreeEnergyPrinciple2017} for reviews). Active inference has more recently gained traction in machine learning and robotics (see \cite{lanillosActiveInferenceRobotics2021,dacostaHowActiveInference2022a,mazzagliaFreeEnergyPrinciple2022} for reviews).

\paragraph{Learning world models:} The key challenge to unlock the usefulness of model-based approaches to AI at scale is to enable agents to learn their model of the world, as current approaches to address this remain limited (but see \cite{heins2025axiomlearningplaygames,fristonPixelsPlanningScalefree2024a,detinguyExploringLearningStructure2024}). Note that this problem is shared by active inference, model-based reinforcement learning and control \cite{pouncyWhatModelModelBased2021}. This structure learning problem\footnote{Here we use the term structure learning synonymously with causal representation learning \cite{scholkopfCausalRepresentationLearning2021,petersElementsCausalInference2017}, or causal model discovery, as is common in the neural and cognitive sciences \cite{gershmanLearningLatentStructure2010}. That is, we use structure learning to denote the joint problem of learning causal relationships and latent representations.} forms the focus of this article: we examine how AI systems and agents might tractably learn models of their data generating process. We examine this problem in detail and discuss the essential roles of core knowledge, information geometry and model reduction, and suggest core structural modules to enable learning a wide range of naturalistic worlds.

\paragraph{Alignment through structure learning:} We then consider AI alignment through the lens of structure learning and active inference. In active inference, the agent's world model supplies the agent's preferences, because behavior simply maximizes the evidence for the model. The thesis we develop is that progress on AI alignment could be made by allowing agents to infer other agents' world models, which contain their preferences. Actions are then mandated to fulfill another's preferences, which corresponds to aligning with the other by taking the other's perspective. This is a subjective notion of alignment that lies beyond formulating safeguards to behavior, which forms the core of many current approaches to alignment. As an illustrative application of these ideas, we mathematically sketch Asimov’s Laws of Robotics in this framework, which prescribe agents that act cautiously to minimize the ill-being of other agents.

\paragraph{Our contribution:} \emph{This article provides a research roadmap toward scalable, aligned AI agents, identifying key research questions and promising directions of travel}. This roadmap comprises three core elements: 1) A commitment to naturalistic approaches to AI that draw from biological intelligence. 2) An integrated framework for agents to autonomously learn world models, enabling scalable intelligence in principle. This synthesizes insights from currently disparate fields---including machine learning, cognitive science, mathematics, and statistics. 3) A conceptual exploration of how these advances might address AI alignment challenges. In summary, this paper provides an integrative perspective and a point of reference for subsequent numerical studies.

\paragraph{Significance of framework:} While the roadmap we present is high-level and conceptual, recent work demonstrates the practical viability of key components. For instance, AXIOM \cite{heins2025axiomlearningplaygames} implements the structure learning principles outlined here (with object-centric active inference models) to master arcade games within 10,000 interaction steps---achieving remarkable sample, computational and parameter efficiency compared to deep reinforcement learning approaches. A similar line of work, theory-based reinforcement learning \cite{tsividisHumanLevelReinforcementLearning2021,pouncyInductiveBiasesTheorybased2022} demonstrates that agents maintaining beliefs over probabilistic programs (encoding causal structure and parameters) can achieve human-level learning efficiency. These implementations illustrate core aspects of this roadmap and demonstrate that naturalistic, structure-learning approaches can deliver both interpretability and performance gains over conventional methods. Our work further suggests that these and related advances could be directly relevant to addressing AI alignment challenges.

\paragraph{Organization of paper:} We set the stage by briefly introducing a `first principles' framework for natural intelligence in Section \ref{sec: fep}. We then discuss learning models of the data generating process (i.e. structure learning and causal representation learning) for static datasets in Section \ref{sec: BSL}, and in an agentic context in Section \ref{sec: SLA}. We then turn to the problem of AI safety and alignment as a potential application of these ideas in Section \ref{sec: alignment}. Finally, we conclude by discussing this naturalistic approach within the broader cognitive science and AI landscape in Section \ref{sec: discussion}.

\section{First principles approach to natural intelligence}
\label{sec: fep}

We aim for approaches to AI that build upon natural intelligence. In what follows, we outline a `first principles' description of intelligent systems that is consistent with the natural sciences, particularly with physical descriptions of the natural world. Specifically, we present a physical theory describing the dynamics of systems that actively interact with their environment.

\paragraph{Notation:} In the following, and unless stated otherwise, we will denote stochastic processes on a finite time interval by lowercase letters and index these stochastic processes by time to denote their (random) value at some timepoint(s). We denote by $P$ their probability distribution.

\begin{figure}
    \centering
    \includegraphics[width=\linewidth]{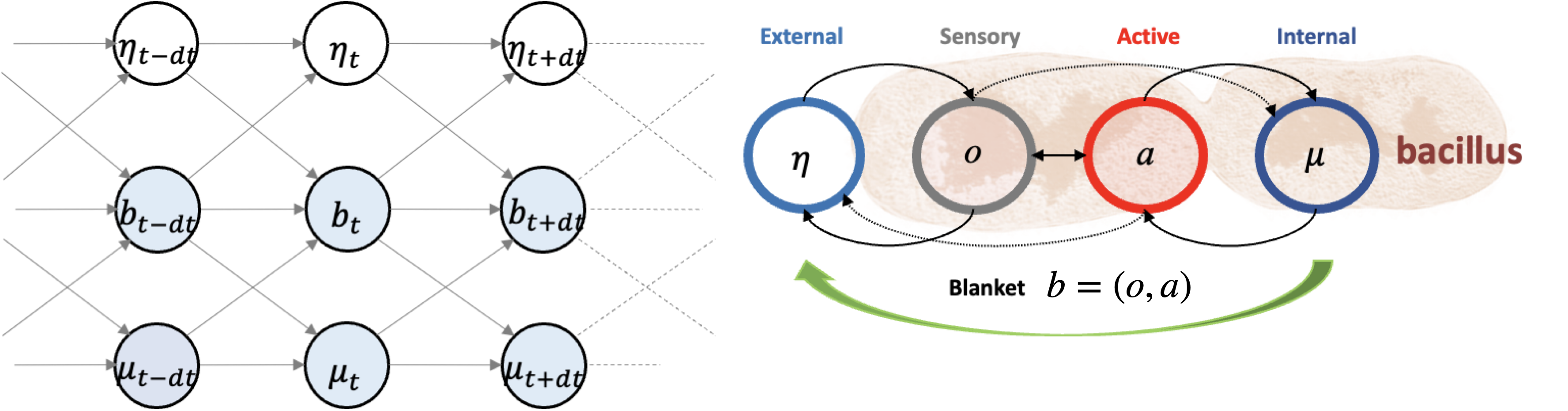}
    \caption{\textbf{A minimal setup.} This figure illustrates our setup: there exists a boundary shielding the internal dynamics of the system from its external dynamics, which persists over some period of time. \textit{Left:} Causal Bayesian network of the interactions between external $\eta$ and internal processes $\mu$ via the boundary process $b$. The boundary is a \textit{Markov blanket} between stochastic processes. In this panel the system is Markovian but it does not need to be in general. \textit{Right:} Causal Bayesian network of the interaction between external, sensory, active and internal processes, where the blanket has been decomposed into sensory and active processes. By definition, internal processes do not directly influence sensory processes and external processes do not directly influence active processes. In many circumstances, internal states parameterize an approximate posterior over external states given blanket states, and evolve consistently with variational inference. This synchronization is represented by the green arrow.}
    \label{fig: MB}
\end{figure}

\paragraph{The setup:} We summarize, under minimal assumptions, the various components of a particle that has internal states, such as an organism or agent (see Figure \ref{fig: MB} for an illustration): Consider the world $x$, comprised of an object of study---such as a particle, organism or agent---and its environment. This partition implies a boundary through which states internal to the particle interact with external states. Thus, the world process $x$ partitions into a process external to the agent $\eta$, a process internal to the agent $\mu$, and a boundary process $b$. Explicitly: $x=(\eta,b,\mu)$. We further decompose the boundary process $b$ into two processes; those that are not directly influenced or caused by the external and internal processes respectively (which may or may not be empty): we call these the active $a$ and sensory $o$ processes respectively $b=(o,a)$. Here, we can interpret the distribution of the world as a generative model for how external processes affect the agent, i.e. Bayes rule
\begin{equation}
\label{eq: Bayes rule}
    \overbrace{P(\eta \mid o,a,\mu)}^{\text{Posterior}}\overbrace{P( o,a,\mu)}^{\text{Evidence}}= \overbrace{P( o,a,\mu\mid\eta )}^{\text{Likelihood}}\overbrace{P( \eta )}^{\text{Prior}}
\end{equation}

\textbf{Maximizing model evidence:} A tautology here is that the most likely internal and active dynamics will maximize the evidence for a generative world model \eqref{eq: Bayes rule}. Precisely, the more likely a trajectory of active and internal processes (given a sensory trajectory), the higher the model evidence---and vice-versa. This is a simple observation that underwrites all that follows: \textit{we can frame internal and active dynamics of things as optimizing one single objective: the evidence for a generative model of the world}. In what follows, we review characterizations of these self-evidencing dynamics in natural systems. The following characterizations assume some functional form to the dynamics of the world, usually (but not limited to) a stochastic differential equation, as these form the basis of a large part of physics, for instance statistical and classical physics, and we aim for a description of natural intelligence  that is consistent with the rest of physics.

\paragraph{Active inference:} Under minimal assumptions, internal states (resp. paths) can be framed as inferring external states (resp. paths) given boundary states (resp. paths) consistently with variational inference in statistics and machine learning \cite{bleiVariationalInferenceReview2017,bishopPatternRecognitionMachine2006}. For example, internal states parameterize probability distributions over external states $\mu_t \mapsto Q_{\mu_t}\left(\eta_t\right)$ such that internal and active states descend (denoted by $\searrow $) variational free energy $\F$ (as usually denoted in statistical physics), which is the negative of the evidence lower bound, or ELBO, that is used in statistics and machine learning:
\begin{equation}
\label{eq: fe principle}
\begin{split}
a_t,\mu_t \searrow \F\left[Q_{\mu_t}, o_t\right]&\triangleq -\operatorname{ELBO}\left[Q_{\mu_t}, o_t\right]\triangleq \E_{Q_{\mu_t}}\left[\log Q_{\mu_t}(\eta_t)-\log P(x_t)\right] \Rightarrow Q_{\mu_t}\left(\eta_t\right) \approx P\left(\eta_t \mid b_t\right).
\end{split}
\end{equation}
This description is known as active inference, as optimizing a free energy or evidence lower bound corresponds to variational Bayesian inference, and active inference extends this by inserting action into the same inferential objective. See \cite{fristonFreeEnergyPrinciple2023a} (resp. \cite{fristonPathIntegralsParticular2023}) for a derivation of \eqref{eq: fe principle} in the case of inference about states (resp. paths), and \cite{dacostaBayesianMechanicsStationary2021a,dacostaTheoryGeneralisedCoordinates2024,fristonStochasticChaosMarkov2021,heinsSparseCouplingMarkov2022} for further details.

A complementary perspective on variational inference is the distribution over internal and active paths. At the scale of classical mechanics, the internal and active paths follow a Boltzmann-Gibbs distribution
\begin{equation}
\label{eq: Gibbs EFE}
    \mu, a \mid d \sim \exp(-\G),\quad \G \triangleq \mathbb E_{P(\eta \mid \mu, a,d)}[\log P(\eta \mid \mu, a,d) -\log P(o ,\eta \mid d)]
\end{equation}
with potential function $\G$ known as \textit{expected free energy} by analogy with the variational free energy over trajectories, and where $d\triangleq d_t\subseteq \{o_{\leq t},a_{\leq t},\mu_{\leq t}\}$ is a potentially non-empty history of data. See \cite{fristonFreeEnergyPrinciple2023a,fristonPathIntegralsParticular2023,barpGeometricMethodsSampling2022a,dacostaActiveInferenceModel2022} for derivations of \eqref{eq: Gibbs EFE}. Interestingly, \eqref{eq: Gibbs EFE} can be related to several information theoretic formulations of intelligent decision-making that predominate in statistics, cognitive science and engineering: see \cite{barpGeometricMethodsSampling2022a,sajidActiveInferenceBayesian2022,fristonSophisticatedInference2021}.

In conclusion, \eqref{eq: fe principle}-\eqref{eq: Gibbs EFE} are two facets of the same coin, in the sense that they are complementary characterizations of the same underlying dynamics. In other words, natural systems may be described as optimizing the model evidence for a generative model of the world, minimizing free energy and pursuing trajectories that minimize expected free energy. 

\subsection{The active inference framework}
\label{sec: aif} 

This theory underpins a normative framework for modeling and simulating the internal and active dynamics of things, such as cognition and behavior, known as active inference \cite{dacostaActiveInferenceDiscrete2020,parrActiveInferenceFree2022,buckleyFreeEnergyPrinciple2017,fristonActionBehaviorFreeenergy2010}. In active inference, internal and active dynamics are taken to maximize the evidence for a generative `world' model $P$ that specifies the interactions between external, sensory and active processes. This is instantiated by numerically minimizing variational free energy \eqref{eq: fe principle} and/or expected free energy \eqref{eq: Gibbs EFE}. In other words, internal and active dynamics are a function of and only of the generative model. The problem for simulating aligned intelligent behavior therefore rests upon choosing the right kind of generative model. This is an open problem which forms the focus of this article.

We outline two features of active inference that will be relevant later and which contextualize it with other approaches to behavior:

 \paragraph{Model evidence guides behavior:} In active inference, the goal of behavior is to maximize the evidence for a generative model of the world. \textit{This means that the agent's generative model of the world describes how things should behave from its perspective and behavior simply fulfills these preferences}. For example, if we consider the cost function for active and internal trajectories, which is the expected free energy, this decomposes into risk and ambiguity, where risk is the KL divergence between predictions and preferences, a prediction error which the agent seeks to minimize:
\begin{equation}
\label{eq: EFE}
\begin{split}
       \operatorname{G}&= \underbrace{\dkl[\overbrace{P(\eta \mid \mu, a,d)}^{\text{predicted}} \mid \overbrace{P(\eta \mid d)}^{\text{preferred}} ]}_{\text{risk}}+\underbrace{\mathbb E_{P(\eta \mid \mu, a,d)}[-\log P(o \mid \eta,d)]}_{\text{ambiguity}} \\
       &\geq -\underbrace{\mathbb{E}_{P(o \mid a ,\mu, d)}[\log \overbrace{P(o \mid d)}^{\text {preferred}}]}_{\text{expected utility}}-\underbrace{\mathbb{E}_{P(o \mid a,\mu,d)}[\dkl[P(\eta \mid o, a,\mu, d) \mid P(\eta \mid a, \mu,d)]]}_{\text{expected information gain}}.
\end{split}
\end{equation}
The dependence of preferences on the data means that these can be inferred---i.e. learned---over time \cite{sajidActiveInferencePreference2022,dacostaActiveInferenceModel2022}.
This decomposition into risk and ambiguity has technical implications for AI safety that we will develop in Section \ref{sec: alignment}.
There are no native reward or utility functions in active inference, but the expected free energy can be connected to reinforcement learning if we interpret log probabilities as reward functions \cite{dacostaRewardMaximizationDiscrete2023,weiValueInformationReward2024}. On this reading, the expected free energy is a conservative bound on expected utility plus expected information gain \cite{sajidActiveInferenceBayesian2022}. 

\paragraph{Drawing the boundary around the agent's brain.} Another feature of active inference is that the agent's body is usually modeled as part of the external process. That is, when modeling intelligent agents like ourselves, the boundary between internal and external is typically drawn around the agent's brain as opposed to around its body. For example, to simulate an arm movement in active inference, the locations of the arms will be part of the external process, the sensory process will be the brain's sensations about the arm's locations, and the actions would be the ways in which the brain can influence these locations \cite{parrComputationalNeurologyMovement2021a}. This contrasts with most reinforcement learning schemes \cite{abelThreeDogmasReinforcement2024}.

\section{Bayesian structure learning}
\label{sec: BSL}

Structure learning, here used synonymously with causal representation learning, is \textit{the problem of learning the mechanisms of cause and consequence in the data generating process }\cite{scholkopfCausalRepresentationLearning2021,gershmanLearningLatentStructure2010}. This is a fundamental problem in causality, cognitive science and artificial intelligence: Indeed, cognitive development can be seen as a structure learning process \cite{tenenbaumHowGrowMind2011,ullmanBayesianModelsConceptual2020} and structure learning may be a way toward human-like artificial intelligence by starting from a child's mind and gradually growing it into an adult mind, as already argued by Turing \cite{turingCOMPUTINGMACHINERYINTELLIGENCE1950}.

\subsection{The problem}

The data generating process is an unknown (causal) Bayesian network $\eta$, with unknown latent variables and causal relationships. The reason for this is fundamental: Bayesian networks are a natural mathematical formalism for accounting for random variables and their causal relationships \cite{pearlCausality2009}: all data generating processes can be expressed as  Bayesian networks (or more generally, probabilistic graphical models), and when they are, their causal mechanisms are made transparent.

A Bayesian network or graphical model $\eta$ (henceforth \textit{model})
has three components \cite{pearlCausality2009}: 1) The causal network $m$, formed of the latent representations (nodes) and their relations of causality (directed edges); 2) The parameters of the causal maps $\theta$ (assuming a functional form to these maps); 3) The latent states $s$ (i.e. the states of the representations).\footnote{The distinction between parameters and states is not common in the Bayesian network literature, however this distinction is standard, and will be useful when dealing with, state-space models for time-series data, as will be appropriate for agents.}
\begin{equation}
\label{eq: model factorisation}
    \eta=(s,\theta,m)
\end{equation}
\textbf{The problem,} we argue below, is to \textit{find a generative model that maximizes the marginal likelihood for the data, where we marginalize over the states, parameters \emph{and structure} of the latent Bayesian network.} In this section, we consider a static dataset $d$ (i.e. offline learning); we will reincorporate dynamics in Section \ref{sec: SLA}.

\subsubsection{Optimizing marginal likelihood...}
\label{sec: optim marginal likelihood}

We want to obtain a generative model $P(d, \eta)$, that maximizes the model evidence $P(d)$ (a.k.a., marginal likelihood) for the data. This formally furnishes a minimal length description for the data \cite{federMaximumEntropySpecial1986,grunwaldMinimumDescriptionLength2007}. The log evidence factorizes into accuracy minus complexity
\begin{equation}
\label{eq: accuracy complexity}
    \log P(d)=\underbrace{\mathbb E_{P(\eta \mid d)}[\log P(d\mid \eta)]}_{\text{Accuracy}}-\underbrace{\operatorname{\dkl}[P(\eta \mid d) \mid P(\eta)]}_{\text{Complexity}}
\end{equation}
where accuracy quantifies how much posterior beliefs accurately fit the data and complexity quantifies how far the posterior moves from the prior. Maximizing accuracy entails maximum likelihood inference, while minimizing complexity enforces a constrained maximum entropy (technically minimum relative entropy) that regularizes the posterior. The complexity can also be seen as a proxy for the computational cost of inference, and via Landauer's principle, energetic cost \cite{landauerIrreversibilityHeatGeneration1961}.\footnote{The evidence lower bound \cite{bishopPatternRecognitionMachine2006,bleiVariationalInferenceReview2017}, a.k.a. variational free energy, factorizes in the same way as \eqref{eq: accuracy complexity}, replacing posteriors with approximate posteriors. In this case the complexity term is a proxy for the computational cost of approximate inference.} In short, optimizing the marginal likelihood with respect to some data yields models that are maximally accurate, but also minimally complex, instantiating a form of Occam's razor. 

\subsubsection{... through a variational bound}

\begin{figure}
    \centering
 \includegraphics[width=0.6\linewidth]{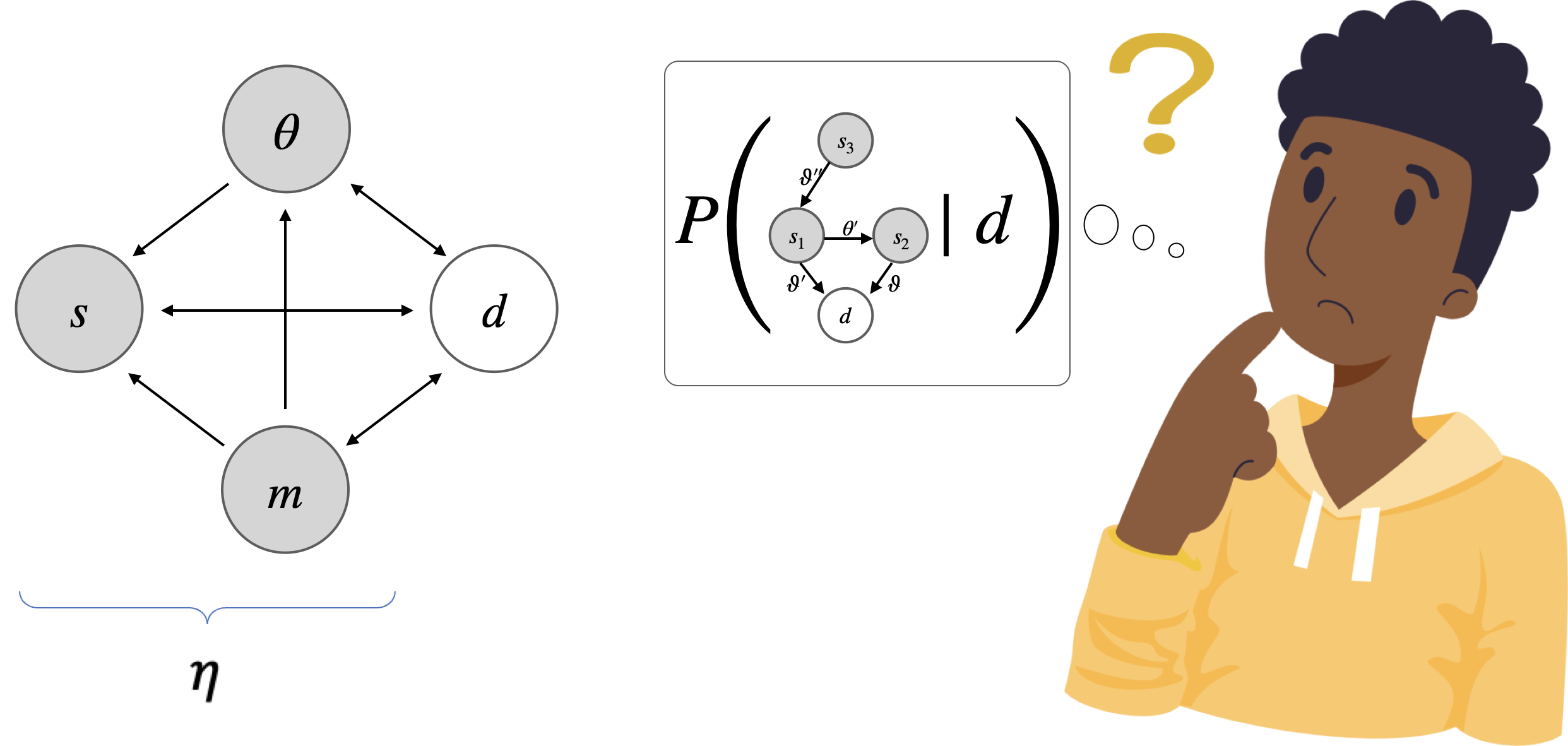}
    \caption{\textbf{Bayesian structure learning.} This figure represents the Bayesian view on structure learning. From data $d$ one must infer the hidden states $s$, parameters $\theta$, and causal network $m$ of the underlying (causal) Bayesian network $\eta$ that describes the data generating process. This means that the number of latent variables and causal structure in the Bayesian network is a random variable that needs to be inferred. The structure learning problem can be solved through hierarchical Bayesian inference, by optimizing an approximate posterior belief over Bayesian networks, which may differ in their latent states, parameters and causal structures. 
    The left panel shows that the variable $m$ causes the parameters $\theta$, and together they cause the states $s$, which together cause the data $d$. The data $d$ may additionally causally influence states, parameters and structure; this will be the case later when part of the data are actions upon the external process (i.e., interventional data \cite{petersElementsCausalInference2017}).}
    \label{fig: CRL}
\end{figure}

Because the marginal likelihood is intractable to compute exactly, we optimize a variational bound: the variational free energy $\F$ or evidence lower bound. In particular this entails approximate Bayesian inference over the latent Bayesian network $\eta$, by optimizing an approximate posterior distribution $Q(\eta)$ over the network structure $m$, parameters $\theta$, and states $s$; see Figure \ref{fig: CRL}. From \eqref{eq: model factorisation}:
\begin{equation}
\label{eq: VI}
\begin{split}
        \F[Q(\eta)]&\triangleq \E_{Q(\eta)}[\log Q(\eta)-\log P(\eta,d)]\\
    &=\underbrace{\dkl[Q(\eta) \mid P(\eta)]}_{\text{Complexity}}-\underbrace{\E_{Q(\eta)}[\log P(d \mid \eta)]}_{\text{Accuracy}}\\
    &=\underbrace{\dkl[Q(\eta)\mid P(\eta\mid d)]}_{\text{Inference}}-\log P(d)\\
    &=\underbrace{\dkl[Q(m)\mid P(m\mid d)]}_{\text{Inference (structure)}}+\E_{Q(m)}\underbrace{\dkl[Q(\theta\mid m)\mid P(\theta\mid m, d)]}_{\text{Inference (parameters)}}\\
&+\E_{Q(m,\theta)}\underbrace{\dkl[Q(s\mid m,\theta)\mid P(s\mid m,\theta, d)]}_{\text{Inference (states)}}-\log P(d)\geq -\underbrace{\log P(d)}_{\text{(log) evidence}}.
\end{split}
\end{equation}
In the last line of \eqref{eq: VI}, we exploit the fact that the approximate posterior distribution factorizes as $Q(m,\theta,s)=Q(s \mid m,\theta)Q(\theta \mid m)Q(m)$ so that it is possible to decompose the problem into hierarchical inference about states, parameters and structure.

\begin{remark}[Encoding uncertainty about structure]
\label{rem: CRL}
    Compare the problems of maximizing the evidence \eqref{eq: accuracy complexity} with the problem of finding the structure with the highest marginal likelihood: i.e. $\arg\max_m P(d\mid m)$. The latter can be seen as doing maximum a posteriori inference (MAP) about structure---i.e. $Q(m)$ is a point mass in \eqref{eq: VI}---with a uniform prior $P(m)$ over structures. This also corresponds to maximizing the likelihood of the structure given the data (i.e. maximum likelihood). However, our prior knowledge about structures is generally not uniform, making the prior $P(m)$ non-uniform. Furthermore, in the finite and even infinite data regime there may be multiple structures with the same likelihood (i.e. \textit{unidentifiability} \cite{rebaneRecoveryCausalPolytrees1987,petersElementsCausalInference2017}), implying that entertaining one single structure is prone to over-fitting. To avoid this, it helps to entertain a richer family of approximate posterior distributions that encode uncertainty about structure in \eqref{eq: VI}.
\end{remark}

Maximizing the marginal likelihood over Bayesian networks by optimizing the variational bound \eqref{eq: VI}
 is a very difficult problem to solve at scale \cite{chickeringLearningBayesianNetworks1996,chickeringLargeSampleLearningBayesian2004}. One of the main intrinsic difficulties lies in the fact that the number of possible causal networks increases super-exponentially in the number of latent variables \cite{weissteinAcyclicDigraph}, hence the space of models that might explain any given dataset \textit{a priori} is huge. In the following, we discuss ways to optimize the variational bound with respect to the prior and approximate posterior, with the aim of producing more scalable methodologies.

\subsection{The prior: model reduction}

The prior $P(\eta)$ should represent the prior state of knowledge about the external world and not overcommit to certain hypotheses \textit{a priori} when they are not directly supported by prior knowledge. For example, it is common to argue that the prior should be the maximum entropy distribution that is consistent with prior knowledge when this is expressed in terms of constraints on that distribution \cite{jaynesPriorProbabilities1968}.

Bayesian model reduction \cite{fristonPostHocBayesian2011,fristonBayesianModelReduction2019,dacostaActiveInferenceDiscrete2020} is an extremely effective computational tool for selecting better priors \textit{after} receiving some data. The idea is to have a collection of prior distributions $P_\lambda(\eta) \triangleq P(\eta \mid \lambda)$ indexed in some set $\lambda \in \Lambda $. Then the model evidence (and posterior) become dependent on $\lambda$ even though the likelihood is fixed
\begin{align*}
    P_{\lambda}(\eta \mid d)=\frac{P(d\mid \eta)P_{\lambda}(\eta)}{P_{\lambda}(d)}.
\end{align*}
Performing variational inference for one base prior $P_{\lambda}(\eta)$ yields the posterior $P_{\lambda}(\eta \mid d)$ (practically, an approximation thereof). It follows that we can score the (relative) model evidence for other parameters $\lambda'$ as a function of the already available $P_{\lambda}(\eta \mid d)$ \cite[Eq 26]{dacostaActiveInferenceDiscrete2020}
\begin{equation}
\label{eq: BMR}
    \frac{ P_{\lambda'}(d)}{ P_{\lambda}(d)}= \mathbb{E}_{P_{\lambda}(\eta \mid d)}\left[\frac{P_{\lambda'}(\eta)}{P_{\lambda}(\eta)}\right].
\end{equation}
The advantage of \eqref{eq: BMR} is that it allows us to find the prior that maximizes the evidence without computing other approximate posteriors (which is the expensive part) yielding
\begin{equation}
\label{eq: maxl lambda}
     \max_{\lambda \in \Lambda} P_\lambda(d).
\end{equation}
Technically, achieving \eqref{eq: maxl lambda} is maximum likelihood inference on the parameter $\lambda $ given the data. To avoid overfitting the data, it is important to employ BMR only after a sufficient amount of data has been acquired.
See Appendix \ref{app: BMR} for a practical summary of BMR.

\subsection{The prior: information geometry}

The space of models has some structure to it---intuitively, a geometry---and this structure should be accounted for in the choice of prior, and in the variational inference problem at hand.

\begin{figure}
    \centering
    \includegraphics[width=\linewidth]{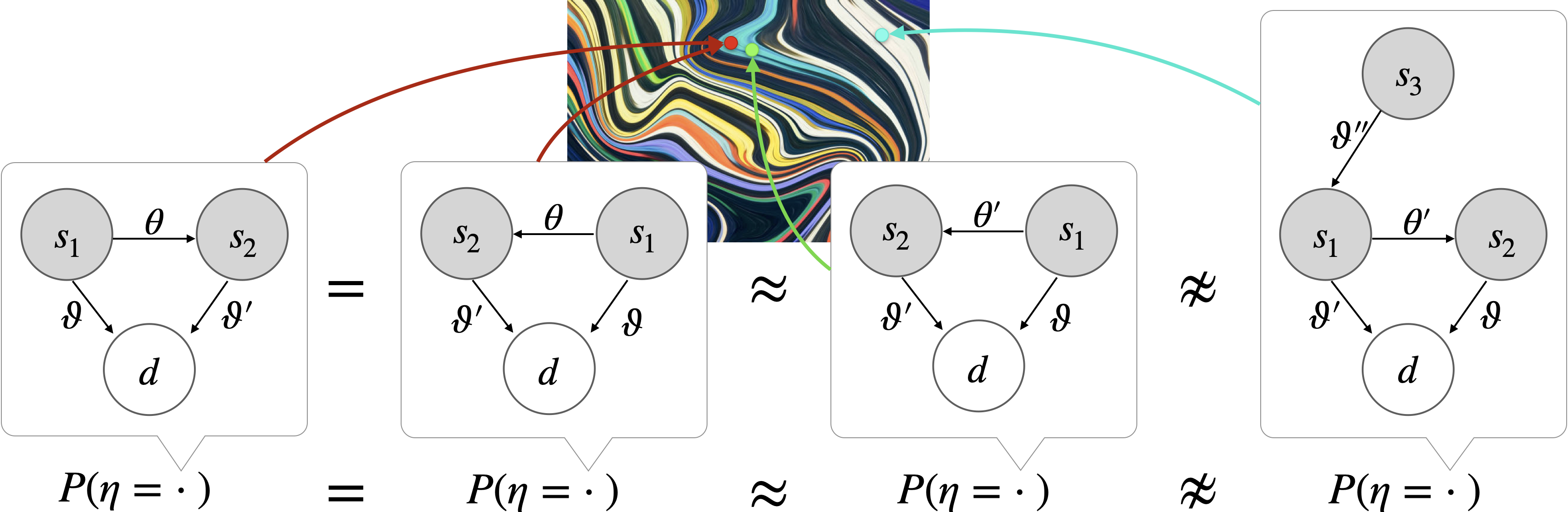}
    \caption{\textbf{Information geometry of models.} \textit{Bottom:} We represent four different models with their states $s$ and parameters $\theta$, which are random variables. Notice that the left-most two express the exact same information (they are the same up to symmetry), hence the prior probability for either of these networks should be the same. The third model is slightly different from the second due to the change of one random variable $\theta$ by another $\theta'$. We would expect these networks to convey relatively similar information, therefore their prior probability should be similar. In contrast, the fourth model is quite different from all three other networks due to the addition of one extra node and causal map. Therefore the prior probability for this network might be very different from all other three. In sum, similarity in the information conveyed implies (local) consistency constraints on the prior over Bayesian networks. \textit{Top:} This illustrates the space of models as a stratified space: a disjoint union of different strata, each of which defines a set of models whose joint space of states and parameters coincide. Here, the bottom three left-most models belong to the same stratum, but the bottom right-most model belongs to a different strata.}
    \label{fig: inf geom}
\end{figure}

Regarding the prior, if two models express the exact same information, they should be assigned the same prior probability, and if they express a similar amount of information they should be assigned a similar prior probability---see the illustration in Figure \ref{fig: inf geom}. Mathematically it seems that there ought to be an information geometry (i.e. a notion of distance) on the space of models that expresses the extent to which two models differ in their information content, and the prior should be continuous in the associated topology (i.e. map similar models to similar probabilities).

An information geometry is induced by a distance or divergence \cite{ayInformationGeometry2017,amariInformationGeometryIts2016}; so what is the natural information distance or divergence on the space of models and what might be ways of tractably implementing this ideal in practice? The difficulty with these questions is that the space of models seems to be a \textit{stratified space}, i.e. a disjoint union of different \textit{strata}
where each strata is a space of probability distributions on the same underlying space; that is, a set of all models with the same joint space of states and parameters. Mathematically, each strata defines a statistical manifold with a well defined information geometry \cite{ayInformationGeometry2017}, but this geometry does not seem to extend to measuring distance between different strata or elements belonging to different strata. This is because the usual notions of information distance, when applied to two Bayesian networks that differ in their joint space of states and parameters yields infinity. In sum, the classical theory of information geometry does not address this problem. Endowing the space of models with a meaningful information distance that can be implemented in practice should be very helpful for structure learning: for affording local consistency constraints on the prior, and for furnishing natural gradients \cite{amariNaturalGradientWorks1998} which supply locally optimal updates to follow during variational inference. 

Looking forward, we should step back and consider the problem of Bayesian inference on \textit{the space of models endowed with an information geometry}, so that models that include the exact same information are identified, and so that \textit{we infer equivalence classes of models up to information equivariance}. For example, the number of symmetries that render two models equivalent increases vastly in the number of latent variables considered. These information invariances are not generally factored into current methods, which means that the model space they face is overwhelmingly larger than it need be. Quotienting by these (and other) invariances should greatly reduce the complexity of the problem---and highly improve model evidence. Furthermore, specifying the prior over equivalence classes has an advantage. Consider that the prior probability of a model equivalence class equals the sum of the prior probabilities for all elements in the class. When specifying prior probabilities over individual models as opposed to equivalence classes, we may have unintended over-counting effects; leading to a higher prior for model equivalence classes with a large number of elements, such as those comprised of models with a high number of latent variables. Practically addressing this issue seems important to scale Bayesian structure learning.



\subsection{The posterior: approximate inference over structures}
\label{sec: approx inf}

\begin{figure}
    \centering
    \includegraphics[width=0.8\linewidth]{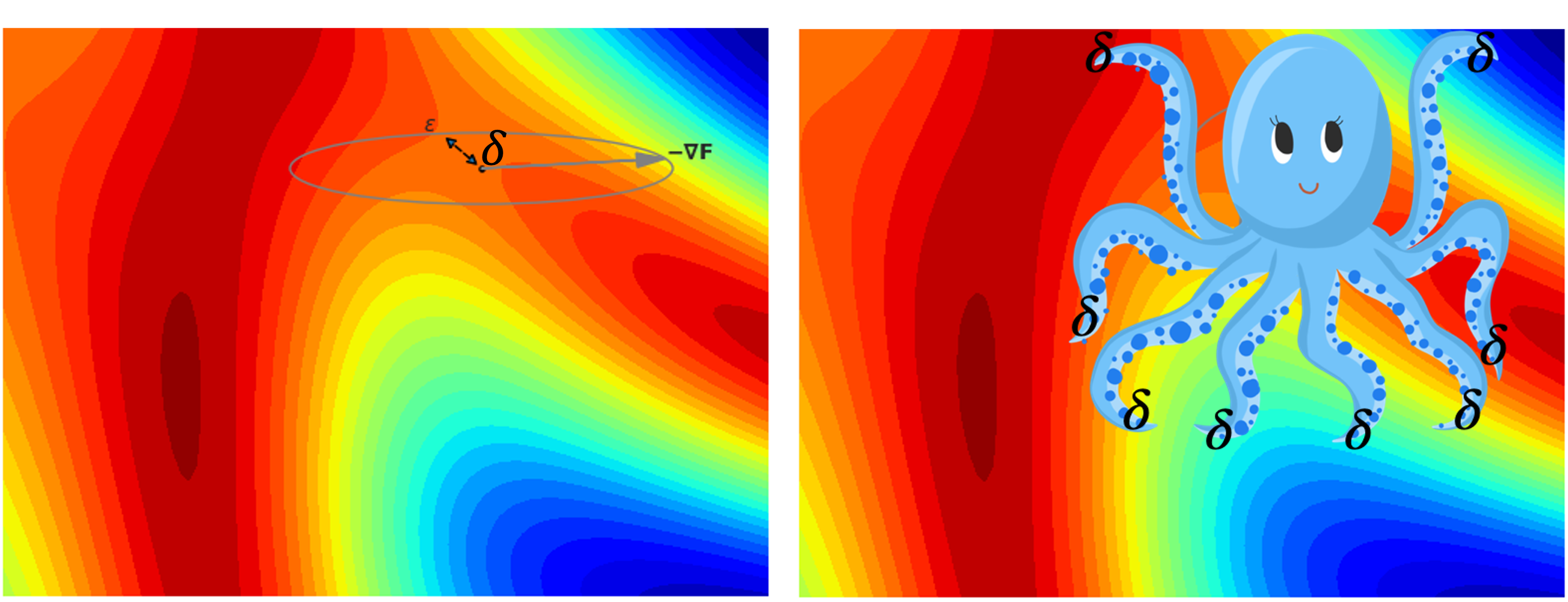}
    \caption{\textbf{Particle variational inference on metric spaces.} These panels show an optimization landscape of a function $\F$ as a two dimensional heat map. \textit{Left:} This shows how we can take gradients of a function $\F$ (e.g. the free energy) as soon as the space of its arguments admits a metric structure (i.e. a notion of distance). Indeed, for any $\e>0$ the gradient of a function $\F$ at an argument (here denoted as a single particle $\delta$) is the direction toward the argument at distance $\e$ from $\delta$ that yields the most descent in  $\F$. Setting $\e>0$ to be small yields the local direction of steepest descent. 
    \textit{Right:} Another approach is (multi-particle) Bayesian optimization, where the Dirac masses are updated through a mixture of exploration of the unknown optimization landscape and exploitation to seek the optimum. This is akin to an Octopus moving in its eco-niche in search of the most favorable locations, where each of its $n$ arms represent the location of one particle.}
    \label{fig: approx inf}
\end{figure}

It remains to optimize the variational bound \eqref{eq: VI} with respect to the approximate posterior $Q(m)$: i.e. variational inference. 
We focus on how to infer structures variationally by optimizing $Q(m)$ to match $P( m \mid d)$ in \eqref{eq: VI}. This is because states and parameter inference---given a structure---is a solved problem in the cases we will consider later \cite{winnVariationalMessagePassing2005,yedidiaConstructingFreeEnergyApproximations2005,murphyLoopyBeliefPropagation1999}. The space of structures $m$ is inherently discrete and thus, the posterior distribution $P(m \mid d)$ is a categorical distribution. This means that the approximate posterior $Q(m)$ must also be categorical. We summarize the representative approaches to structural inference based on the parameterization of the approximate posterior (see Figure \ref{fig: approx inf} for an illustration):

\begin{enumerate}
    \item \textbf{Particle approximate posterior} $Q\left(m \mid n, \lambda_i, m_i\right)=\sum_{i=1}^n \lambda_i \delta_{m_i}(m)$: This is when the variational inference method entertains a (typically small) number $n \geq 1$ of structures $m_i$, which are optimized to capture the modes of the posterior distribution, and whose respective posterior probabilities $\mu_i$ are optimized accordingly. In this setting, we can optimize the structures being considered by making small or large updates:
    \begin{itemize}
        \item Local updates: 
        \begin{enumerate}
            \item \textbf{Markov chain Monte-Carlo (MCMC)} approaches run a stochastic process on the space of structures to sample the true posterior. Samples are produced sequentially by the process according to some stochastic rule (e.g. adding nodes in a Bayesian network with some probability). The process is ensured to converge to the target in distribution through some consistency procedure like Metropolis-Hastings \cite{madiganBayesianGraphicalModels1995,giudiciImprovingMarkovChain2003}, and may be optimized in various ways to increase the speed of convergence \cite{eatonBayesianStructureLearning2012,grzegorczykImprovingStructureMCMC2008,kuipersPartitionMCMCInference2017}.
            \item \textbf{Constrained continuous optimization} approaches embed the space of structures as a constraint set in a larger continuous space \cite{zhengDAGsNOTEARS2018,yuDAGGNNDAGStructure2019,lorchDiBSDifferentiableBayesian2021}, thereby finessing the complexities of discrete space variational inference by allowing the use of mature toolboxes from continuous particle optimization, e.g. \cite{liuSteinVariationalGradient2016}, to perform the inference.
        \end{enumerate}
        \item Global updates:
        \begin{enumerate}
            \item \textbf{Discrete particle variational inference} is a variational inference procedure on discrete spaces where structures are updated through a conjugate free energy descent \cite{saeediVariationalParticleApproximations2017}.
            \item \textbf{Bayesian optimization} observes that the free energy of structural inference is an expensive function to evaluate, and makes global updates to each particle through Bayesian optimization \cite{mockusBayesianApproachGlobal1989}. This rests upon having a generative model of the free energy landscape, which is sometimes referred to as meta-modeling. The simplest scheme we envisage is to have a (multivariate Gaussian) prior over the discrete space of structures encoding the \textit{a priori} goodness of each structure; that is, the free energy minimum of each given structure. We can then use an acquisition function (such as expected free energy) to select new structures to evaluate, and once we have committed to some structure, we can infer its parameters and states by following free energy gradients. Crucially, as we evaluate multiple structures, we could learn the covariance of the Gaussian, so that we can empirically learn similarities between structures. This would give us empirical insight on the information geometry between structures, and the resulting covariance could be used as an empirical prior for more efficient Bayesian optimization down the line. One can call this approach an Octopus search because it palpates the free energy landscape with its $n$-arms (the Dirac particles) through a mixture of exploration and exploitation (c.f., unscented filtering in nonlinear estimation).
        \end{enumerate}
    \end{itemize}
    \item \textbf{Fully categorical approximate posterior} $Q(m \mid \lambda)=\operatorname{Cat}(m \mid \lambda)$: This is where the distribution being parameterized is a full categorical distribution, so that we entertain beliefs about the plausibility of a potentially large number of structures, as large as the number of structures being considered in the prior. For this, there are two main approaches:
\begin{enumerate}
    \item \textbf{Amortized inference:} Trains a neural network to predict the variational parameters $\mu$ based on the data \cite{lorchAmortizedInferenceCausal2022}.
    \item \textbf{Generative flow networks (GFlowNets):} are a generic tool for sampling-based approximate inference over discrete compositional spaces, such as the space of models \cite{deleuBayesianStructureLearning2022,deleuJointBayesianInference2023,nishikawa-toomeyBayesianLearningCausal2022}. Technically, this is a state-action policy whose states are structures. For example, actions are adding or removing nodes to a model. The policy implicitly encodes the approximate posterior: the approximate posterior $Q(m)$ is the distribution of its terminal state, so that we can sample from the posterior by running the policy. GFlowNets are trained (implicitly) so that the variational posterior minimizes variational free energy \cite{malkinGFlowNetsVariationalInference2022}.
\end{enumerate}
\end{enumerate}

These methods come with distinct features and trade-offs. For example, methods expressing a full categorical distribution are much more expressive and provide a more accurate solution to the Bayesian inference problem; but they may be slower to train. An important desideratum for a structural inference method is to provide an accurate account of structural uncertainty by finding the multiple modes in the target distribution $P(m \mid d)$ (cf. \cref{rem: CRL}); MCMC approaches are notoriously limited for this while GFlowNets and Bayesian optimization may compare favorably. Compared to the other methods, fully categorical and constrained continuous optimization approaches deal with a purely continuous optimization problem, which facilitates variational inference by furnishing gradients, but which makes inference prone to local minima (other methods are not exempt). Future work, we hope, will comprehensively quantify the features and trade-offs of these approaches. For structure learning agents, an important desideratum is the ability to perform fast inference in real-time.

There are many issues and speculations that follow from this classification, and which we hope, could be addressed in future work: which of these schemes, if any, are biologically plausible? And what may be most apt to account for structure learning in the human brain? How can any of these schemes be augmented with information geometric considerations to make them faster (by using natural gradients) and more scalable (by factoring information invariances)? Is the ability to infer from time-series data online (a feature of real agents) characteristic of particle approximate posteriors? In regard to the first two questions, we note that the brain can only store in short-term memory---and attend to---a few objects simultaneously at any given time \cite{cowanMagicalNumberShortterm2001,spelkeCoreKnowledge2007}. Might this be indirect evidence for the hypothesis that the brain could only consider a few competing causal hypotheses about the world, and thus be implicitly encoding a particle approximate posterior with a few particles?

\section{Structure learning agents}
\label{sec: SLA}

\begin{figure}
    \centering    \includegraphics[width=0.8\linewidth]{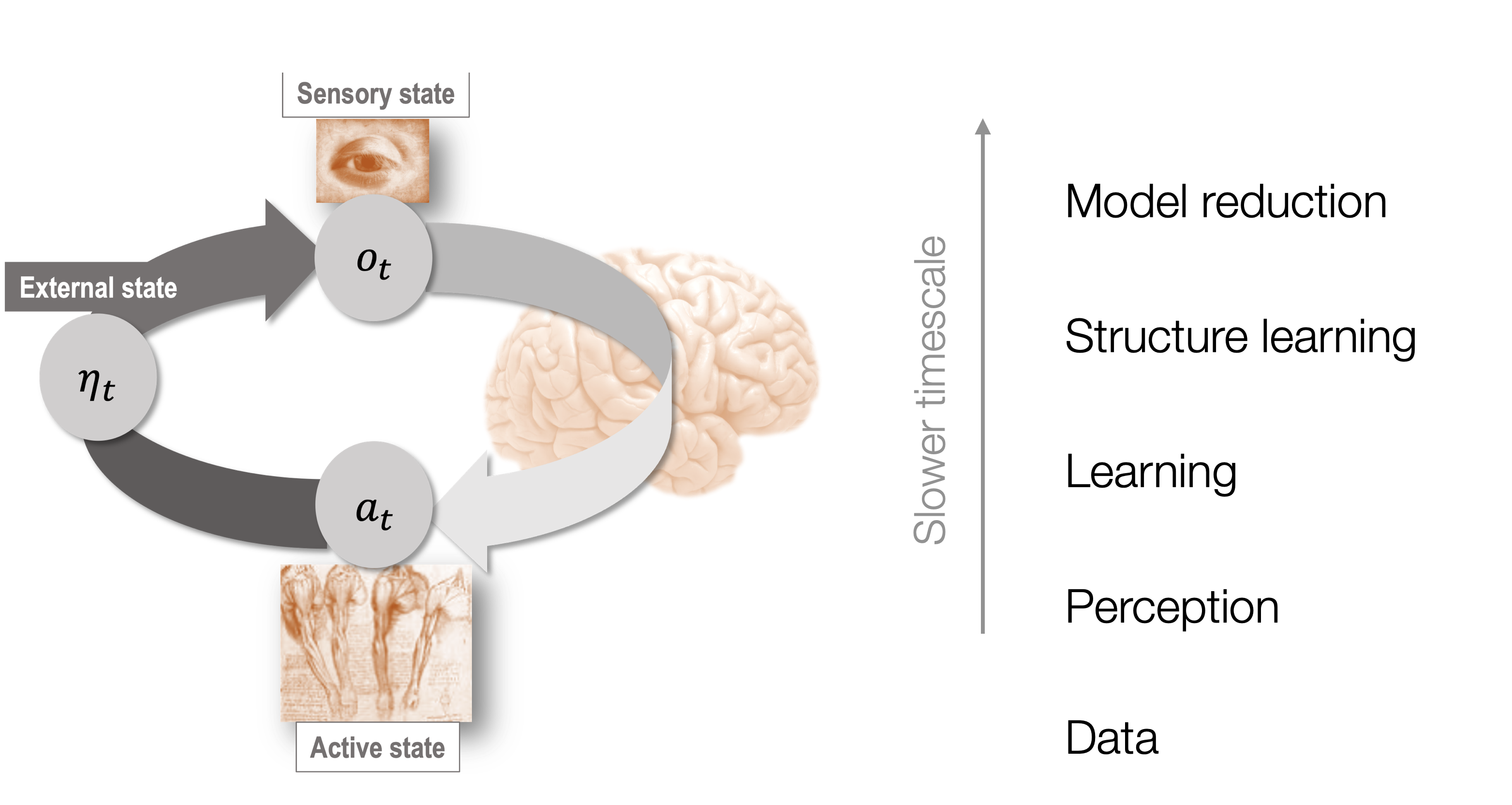}
    \caption{\textbf{Structure learning agents.} This figure summarizes the various processes underlying agents that learn causal structure. \textit{Left:} The external state of the environment causes a sensory state, which is processed by the agent resulting in a choice of active state, which influences the next external state---and the perception action cycle repeats. \textit{Right:} Agents have access to an incoming stream of data, and use this for perception, learning, structure learning and finally model reduction, which have a technical meaning here: namely, inferring the states, parameters, and structure of models, respectively. These processes unfold at slower and slower timescales: this is necessary since accurate inference about e.g. causal structure, requires many more data points than inference about parameters or states, under a causal structure.}
    \label{fig: STL-AIF}
\end{figure}

We now turn to discussing agents that learn the causal structure of the world. The agentic setup is illustrated in Figure \ref{fig: STL-AIF} (left panel): the agent is in dynamic exchange with the external process, whereby the current external state $\eta_t$ yields an observation $o_t$, then the agent takes an action $a_t$, which influences the external process etc; and the perception-action cycle repeats. Compared to Section \ref{sec: BSL},
the agent has access to an incoming stream of (interventional) data $t \mapsto d(t)\triangleq d$ comprised of past sensations and actions $d \subseteq \{o_{\leq t}, a_{\leq t}\}$, that is continuously updated at each cycle.

\subsection{Model-based planning and multi-scale inference}

Following Section \ref{sec: fep}, we propose to investigate this problem through the lens of active inference. Practically, this implies a commitment to model-based planning and multi-scale inference.

\paragraph{Model-based planning:} The agent possesses a generative model about the latent states, parameters, and causal structure describing the world (we will see examples later). It uses this model for planning, by optimizing an objective combining explorative and exploitative drives, such as the expected free energy \eqref{eq: EFE}. 

\paragraph{Multi-scale inference:} The defining feature of Bayesian approaches to behavior is inferring the external process $\eta$ from the data $d$. This involves approximating posterior beliefs such as $P(m,\theta,s\mid d)$ about (the past, present and future) structure, parameters and states of the world. This may be solved variationally \eqref{eq: VI} by updating an approximate posterior distribution $Q(m, \theta, s)$ to match incoming data. As we have seen in \eqref{eq: VI}, this inference can be hierarchically decomposed by inferring states $Q(s\mid m,\theta)$ (i.e. \textit{perception}), then parameters $Q(\theta\mid m)$ (i.e. \textit{learning}) and then causal structure $Q( m)$ (i.e. \textit{structure learning}). Additionally, the agent may engage in Bayesian model reduction to simplify its model of the world.

These inferential processes may operate at different time-scales: perception faster than learning, which is faster than structure learning, which is faster than model reduction. This is because more data is needed for accurate learning than for perception, and even more so for accurate structure learning and ensuing model reduction. There is empirical evidence that the brain complies with this separation of time-scales: perception by neuronal populations is plausibly encoded in their firing rates---which are fast processes---while learning is encoded in the modulation of neural connection strengths (i.e. Hebbian plasticity) that fluctuate much more slowly \cite{isomuraCanonicalNeuralNetworks2022,isomuraExperimentalValidationFreeenergy2023,dacostaNeuralDynamicsActive2021}. Could it be that causal structure is encoded in the functional connectivity between neural populations, and be updated even more slowly? Model reduction can plausibly be interpreted as pruning connections in or between neuronal populations \cite{fristonActiveInferenceCuriosity2017}, as is seen to happen during development and throughout life (e.g., during the sleep-wake cycle). In physics, processes operating at different scales are known as multi-scale processes \cite{pavliotisMultiscaleMethodsAveraging2010}.

To simulate this multi-scale inferential process in practice, one would set the learning rates in the optimization of $Q(m)$ to be much lower than $Q(\theta \mid m)$, which are much lower than $Q(s \mid m,\theta)$. For convenience, what is commonly done in practice is inferring states after every new observation, inferring parameters after every small batch of observations \cite{dacostaActiveInferenceDiscrete2020}, and inferring structure after every bigger batch of observations---and reducing the model after even larger batches. Specifying the respective batch sizes corresponds to specifying the relative timescales of the different inferential processes. In physics, this corresponds to an adiabatic approximation of a multi-scale process \cite{pavliotisMultiscaleMethodsAveraging2010}. See the summary in Figure \ref{fig: STL-AIF} (right panel).

\subsection{Related work}

One very related line of work is theory-based reinforcement learning \cite{tsividisHumanLevelReinforcementLearning2021,pouncyInductiveBiasesTheorybased2022,tomovNeuralArchitectureTheorybased2023}. In a foundational paper \cite{tsividisHumanLevelReinforcementLearning2021}, an agent maintains beliefs about probabilistic programs, which implicitly encode the causal structure, parameters, and states of the world. The agent then optimizes expected utility plus information gain to select the next action (note the similarity with \eqref{eq: EFE}). The authors deployed this architecture in a suite of simplified Atari games, and found that not only did their agent achieve human learning efficiency across the games (after comparing with data from human participants), but the agents' learning trajectories were also relatively similar to that of humans. This work serves as a proof of concept that combining inference about the structure of the world with model-based planning---leveraging both exploration and exploitation---can achieve human-level sample efficiency and performance and relatively human-like behavior.

Current active inference schemes engage in multi-scale perception, learning, structure learning and model reduction \cite{dacostaActiveInferenceDiscrete2020,parrActiveInferenceFree2022,wauthierSleepModelReduction2020,fristonActiveInferenceCuriosity2017}. Structure learning active inference agents is an area of active research and current schemes do hold beliefs about more than one alternative structure \cite{fristonActiveInferenceCuriosity2017,smithActiveInferenceApproach2020,fristonSupervisedStructureLearning2023,detinguyExploringLearningStructure2024,fristonPixelsPlanningScalefree2024}.

\subsection{Refining the search space of possible structures}

Building agents that scalably learn causal models of the world are a relatively open challenge \cite{scholkopfCausalRepresentationLearning2021}. Perhaps the main difficulty is the explosion of the search space of possible structures that might explain more and more complex worlds \cite{weissteinAcyclicDigraph}. To illustrate the problem, consider the above theory-based reinforcement learning work \cite{tsividisHumanLevelReinforcementLearning2021}. The search space of explanatory hypotheses about the world their agents consider is the whole set of programs (up to a certain length) that can be generated from the code generating the computer program that generates the data. This is an extremely large search space even for the simplified Atari environments their agents face, and a feat of this work is that structural inference is made tractable even so; however, this approach is obviously limited in its scalability: 1) in more complex environments, the space of programs that can be generated from the code grammar generating the environment may be far too large to be searchable, 2) in general the modeler does not know the generative process and cannot form a space of candidate explanations that contains the data generating process. We now investigate ways to address these shortcomings by respectively considering core knowledge priors and universal generative models.

\subsubsection{Core knowledge priors}

Core knowledge represents prior knowledge about the external world that would be valid across any world that the agent could be born in. As much as possible, this core knowledge should be reflected in the prior probability for potential model explanations for the world, to reduce the search space of possible explanations. For agents operating in naturalistic worlds, core knowledge may include an intuitive understanding of physics, e.g. statements like ``objects cannot interact at a distance but agents can'', and many others \cite{spelkeCoreKnowledge2007,spelkeWhatBabiesKnow2022}. 

Evolution has carved this kind of core knowledge into our genome so that humans and animal newborns are born with rich prior knowledge about the world. For example, human infants are endowed with at least seven rich systems of core knowledge about objects, places, agents, numbers, geometry, social groups, and others' mental states \cite{spelkeCoreKnowledge2007,spelkeWhatBabiesKnow2022}. These are shared by humans across ages and cultures and sometimes across several animal species \cite{spelkeCoreKnowledge2007}. One can think of the process by which evolution has learned this kind of prior knowledge as a process of evidence maximization on an evolutionary timescale \cite{fristonVariationalSynthesisEvolutionary2023}.

This `common sense' prior knowledge vastly improves the evidence for an agent's model of the world. Core knowledge avoids compromising model accuracy by precluding overly specific assumptions about the natural world---and vastly reduces model complexity by restricting the search space of explanations for the world. This knowledge greatly aids structure learning: core knowledge provides a valid carving of the world into distinct classes of things (such as objects or agents) with distinctive properties, rather than leaving this as structure to be learned. Through this, core knowledge dramatically speeds up inference and learning; for instance, if two things seemingly interact at a distance, then it can be inferred with certainty that at least one of them is an agent. 

Reverse-engineering human and animal systems core knowledge into priors over models or probabilistic programs is an ambitious and ongoing research effort \cite{spelkeWhatBabiesKnow2022,pouncyInductiveBiasesTheorybased2022}. Follow up work on theory-based reinforcement learning encoded core knowledge into soft constraints on the types of programs that might explain a given (Atari) world, and found that the agents followed more human-like learning trajectories with core knowledge than without such inductive biases \cite{pouncyInductiveBiasesTheorybased2022}. In more complex worlds, we hypothesize that core knowledge priors become absolutely essential to learn with any efficiency.

Core knowledge therefore constitutes knowledge that is valid across any naturalistic world, which translates into significant constraints on the prior over models. The prior over models as explanations for the world is then constrained by the consistency with the underlying information geometry (i.e. local constraints) and core knowledge constraints (i.e. non-local constraints).\footnote{The maximal entropy probability distribution that is consistent with an underlying geometry and some constraints is a Gibbs measure on the underlying metric space \cite[Chapter 9]{lasotaChaosFractalsNoise1994}.}

\subsubsection{Toward universal, interpretable, agentic generative models}

A fundamental question is what might be a `universal' set of primitives and compositional rules to produce a space of models as potential explanations for the world that is both \cite{dacostaUniversalInterpretableWorld2024}:
\begin{enumerate}
    \item \label{item: expressive}\textbf{Sufficiently expressive} to be able to approximately express any kind of naturalistic, dynamic interactions between agent and environment.
    \item \label{item: coarse}\textbf{Sufficiently coarse} so that inference on this space can be made computationally tractable.
\end{enumerate}
Furthermore, each model in this space should be:
\begin{enumerate}
\setcounter{enumi}{2}
    \item \label{item: interpretable}\textbf{Interpretable} so that the agent's understanding and ensuing behavior can be easily understood from the model it entertains.
    \item \label{item: fast}\textbf{Supportive of fast action, perception and learning}.
\end{enumerate}

\begin{definition}[Universal generative model]
    We call a space of models satisfying requirements \ref{item: expressive}-\ref{item: fast} a \emph{universal space of models}. A generative model based on a universal space of models is therefore apt to causally explain any kind of naturalistic world; we will call this a \emph{universal generative model}.
\end{definition}

There is already a tension between requirements \ref{item: expressive} and \ref{item: coarse} and a significant difficulty lies in balancing these requirements. Asking what a universal space of models might look like, we first consider the existing literature: Spaces of \textit{probabilistic programs }are easily made extremely expressive, but it is not clear how to do so while keeping them coarse enough for inference to remain tractable. Probabilistic programs are not always easily interpretable, and, barring specific assumptions, do not support efficient perception and learning, as Bayesian inference over states and parameters may require sampling. One example of probabilistic programs that might satisfy these requirements---to a first approximation---are \textit{hierarchical discrete and continuous state partially observed Markov decision processes} (POMDPs) \cite{astromOptimalControlMarkov1965a,bartoReinforcementLearningIntroduction1992}. Indeed, it has been shown that dynamic models with continuous random variables interacting across time are capable of performing Turing-complete computation~\cite{brule2016computational}. Moreover, it is striking that nearly all modeling work in active inference, which spans nearly two decades, employed models that are built by hierarchically stacking these two types of layers \cite{dacostaActiveInferenceDiscrete2020,parrActiveInferenceFree2022,smithStepbystepTutorialActive2022,lanillosActiveInferenceRobotics2021}. This might be a bias, but it nevertheless indicates that this space of models is very expressive in being able to reproduce a wide variety of behavioral simulations and empirical data. Importantly, these networks support fast action, perception and learning where inference about states and parameters is implemented with fast variational inference procedures \cite{dacostaActiveInferenceDiscrete2020,parrNeuronalMessagePassing2019,heinsCollectiveBehaviorSurprise2024,parrActiveInferenceFree2022,fristonActionBehaviorFreeenergy2010}, which have a degree of biological plausibility in being able to reproduce a wide range of features from real neural dynamics, e.g. \cite{fristonHierarchicalModelsBrain2008,fristonActiveInferenceProcess2017,isomuraCanonicalNeuralNetworks2022,isomuraExperimentalValidationFreeenergy2023}. Barring the use of neural networks for expressing non-linearities in these layers \cite{mazzagliaFreeEnergyPrinciple2022}, each of the layers furnishes an interpretable model of dynamics.

\subsubsection{Expressivity in terms of stochastic processes}
\label{sec: expressivity stochastic processes}

From this, we might envision a set of basic structural modules satisfying requirements \ref{item: interpretable} and \ref{item: fast} that can be assembled hierarchically to express a wide range of dynamic agent-environment interactions. Here, we pursue this line of thought by describing two building blocks that can be combined to express a large class of stochastic processes on both discrete and continuous states \cite{dacostaUniversalInterpretableWorld2024}.

\begin{figure}
    \centering
    \includegraphics[width=\linewidth]{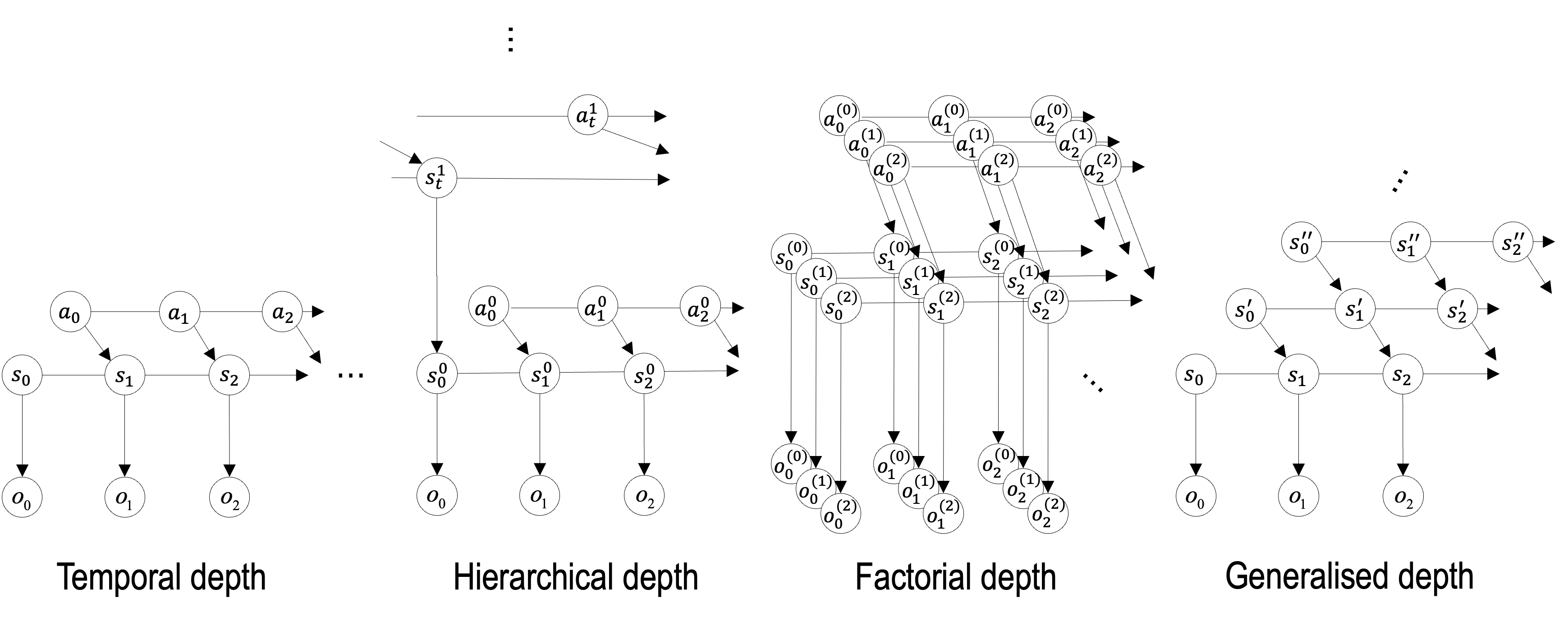}
    \caption{\textbf{Space of discrete-state Bayesian networks.} This figure summarizes an expressive space of discrete state-space models. The basic module is a partially observed Markov decision process (POMDP). This module can have an arbitrary finite temporary horizon (i.e. temporal depth). They can be stacked atop each other finitely many times (i.e. hierarchical depth), thereby expressing multi-scale semi-Markovian latent dynamics. We can specify multiple co-evolving factors (i.e. factorial depth; e.g. position and color of an object). In any given layer, a number of auxiliary (i.e. generalized) states can be added, accounting for velocity, acceleration, and higher orders of motion in the latent states (i.e. generalized depth) to express semi-Markovian processes \cite{fristonSupervisedStructureLearning2023}. In each of these layers, the highest generalized states may or may not be actions denoted by $a$ (cf. controllable states \cite{fristonSupervisedStructureLearning2023}) while all other states are uncontrollable (i.e. part of the environment) denoted by $s$.
    The controllable and uncontrollable states cause observations denoted by $o$. The parameters of these Bayesian networks (not shown) correspond to the parameters of the causal maps as well as the initial distributions over states \cite{dacostaActiveInferenceDiscrete2020}. These parameters as well as the graphical structure need to be inferred from data (e.g. past actions and observations). Please see \cite{fristonSupervisedStructureLearning2023,fristonActiveInferenceIntentional2023a} for more details on temporal, hierarchical and factorial depth---and examples of controllable versus uncontrollable states.}
    \label{fig: discrete}
\end{figure}

\paragraph{Discrete dynamics:} Markov processes are a fairly ubiquitous class of stochastic processes \cite{hairerMarkovProcesses2020}. All Markov processes on discrete states have simple transition dynamics that are given by linear algebra. 
When these transitions also depend on actions, we obtain a Markov decision process. When states are partially observed and observations depend only on the current latent state, we obtain POMDPs. We can add auxiliary latent states to those POMDPs \cite{fristonActiveInferenceIntentional2023a} (i.e. the equivalent of momentum, acceleration etc) to account for the effect of memory in the system, producing semi-Markovian POMDPs. Lastly, we can stack these layers hierarchically to express multi-scale semi-Markovian processes. In summary, extended discrete POMDPs hierarchically compose a very general class of models for agent-environment interactions on discrete states. See Figure \ref{fig: discrete} for a graphical representation of discrete POMDPs and their various degrees of freedom.

\paragraph{Continuous dynamics:} For expressing continuous dynamics, the situation is somewhat more involved. Repeating the construction from discrete state-spaces seems hardly possible because continuous-space Markov processes are given by linear operators in infinite (as opposed to finite) dimensional spaces \cite{bakryAnalysisGeometryMarkov2014}. A working alternative is to restrict ourselves to a more manageable but still very expressive class of processes. We can consider continuous POMDPs with latent dynamics given by stochastic differential equations (SDEs), which is another very expressive class of stochastic processes. Note that the behavior of natural agents is characterized by non-linear dynamics that break detailed balance and colored noise\footnote{Colored noise refers to random fluctuations affecting the motion that have a non-trivial autocovariance function. This means that these random fluctuations have a degree of memory (i.e. non-Markovian); intuitive examples include fluctuations in the wind or in the ocean.}\cite{fristonWhatOptimalMotor2011,parrComputationalNeurologyMovement2021a,fristonActionUnderstandingActive2011,lynnBrokenDetailedBalance2021}, and under active inference, these dynamics must be included in the model because the agent's body is typically modeled as part of the external process (cf. Section \ref{sec: aif}) \cite{fristonActionBehaviorFreeenergy2010}. 
Luckily, there is a remarkably expressive class of SDEs supporting non-linearities, colored noise and broken detailed balance---that is, many times differentiable stochastic differential equations \cite{dacostaTheoryGeneralisedCoordinates2024}---for which POMDPs with these latent dynamics support fast and biologically plausible update rules for action, perception and learning \cite{dacostaTheoryGeneralisedCoordinates2024,heinsCollectiveBehaviorSurprise2024,parrActiveInferenceFree2022,fristonHierarchicalModelsBrain2008}.
These continuous POMDP units yield a very expressive space of continuous-state Bayesian networks by varying the temporal, hierarchical, factorial and generalized depth as in Figure \ref{fig: discrete}.

\begin{figure}
    \centering
    \includegraphics[width=0.6\linewidth]{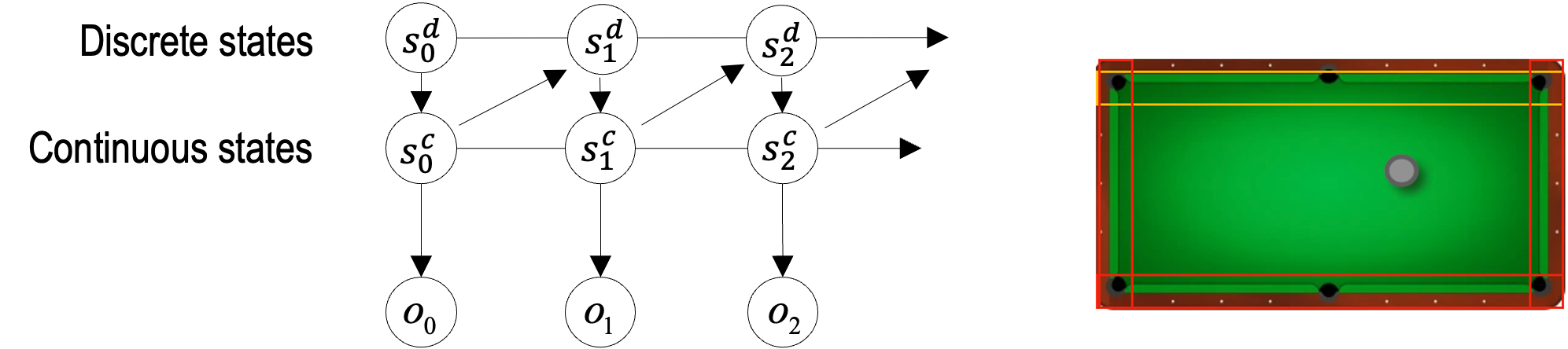}
    \caption{\textbf{Recurrent switching linear dynamical systems (rsLDS).} 
    This figure illustrates rsLDS as a way of parameterizing fairly arbitrary stochastic differential equations (SDEs) as a switching linear SDE. Consider for example a generative model used to play pool. This must be able to express the non-linear trajectories in a game of pool. The behavior of a ball on a pool table depends on whether the ball is by one of the four walls (i.e. bouncing), or not. In each of these sectors the dynamic of the ball is captured by a simple dynamical equation. The rsLDS \cite{lindermanBayesianLearningInference2017} is a simple generative model that is able to express these kinds of trajectories, where a (switching) discrete state expresses in which of these sectors the ball is. This discrete state furnishes a linear drift and volatility to the linear SDE describing the continuous motion of the ball (cf. continuous states), which yields continuous observations. Note the so-called recurrent connection (i.e. causal map) from continuous to discrete states; this connection enables the continuous dynamic to influence the discrete switching: if the continuous trajectory of the ball collisions with a wall the discrete state switches so that the continuous dynamic of the ball changes course. The rsLDS layer may be supplemented with active states, acting on the discrete latent states, thereby emulating a continuous partially observable Markov decision process. This continuous POMDP can easily be extended by varying the temporal, hierarchical, factorial and generalized depth as in Figure \ref{fig: discrete}, furnishing a generic model of continuous dynamics. Please see \cite{lindermanBayesianLearningInference2017} for more details on the rsLDS architecture.}
    \label{fig: rslds}
\end{figure}

One important challenge remains: to parameterize the non-linearities in continuous POMDPs (e.g. flows of SDEs) without sacrificing interpretability, and learn these parameterisations from data. A promising approach is to express non-linear SDEs with recurrent switching linear dynamical systems (rsLDS; see Figure \ref{fig: rslds}) \cite{lindermanBayesianLearningInference2017}; that is, switching mixtures of linear SDEs, because one could use a very fine grained piece-wise linear approximation to recover arbitrary non-linearities, as necessary.
The advantage of using switching linear SDEs is that they are interpretable and afford relatively scalable exact Bayesian inference \cite{lindermanBayesianLearningInference2017}.\footnote{An alternative way of inferring non-linearities would be by parameterising the non-linearities with Bayesian neural networks (e.g. neural SDEs \cite{kidgerNeuralDifferentialEquations2022}), but this would likely hinder interpretability.}
However, rsLDS architectures are restricted to approximating the dynamics of non-linear diffusion processes discretized with an Euler scheme \cite{lindermanBayesianLearningInference2017}, which do not feature colored noise (by definition). Looking forward, it seems apt to extend the rsLDS architecture to express colored-noise SDEs, perhaps by combining it with the machinery of generalized coordinates \cite{dacostaTheoryGeneralisedCoordinates2024}. In particular, this would entail introducing generalized depth into the rsLDS layer. This should furnish an expressive and searchable class of models for expressing continuous-state dynamics satisfying the basic requirements \ref{item: interpretable}-\ref{item: fast}.

\paragraph{Hierarchical mixed dynamics:}
Stacking hierarchies of discrete layers atop hierarchies of continuous layers yields mixed generative models that can express rich non-linearities and dynamics at several levels of abstraction. Despite the absence of traditional neural networks here, these hierarchies form a network, where the layers are discrete and continuous POMDPs and the computations are of efficient approximate Bayesian inference. Hierarchies of these layers may be interpretable as they represent nested processes operating at different timescales. These hierarchical structures are compatible with views of the brain as entertaining discrete-state, low-dimensional abstract dynamics that condition the high-dimensional continuous representations closer to sensory input  \cite{parrDiscreteContinuousBrain2018,fristonGraphicalBrainBelief2017}.

\subsection{Generative models for structure learning agents}

Now that we have seen a space of models that may be apt to describe the dynamical structure of a wide-range of worlds, we return to the generative models that agents might employ to infer this structure.

\begin{figure}
    \centering
    \includegraphics[width=\linewidth]{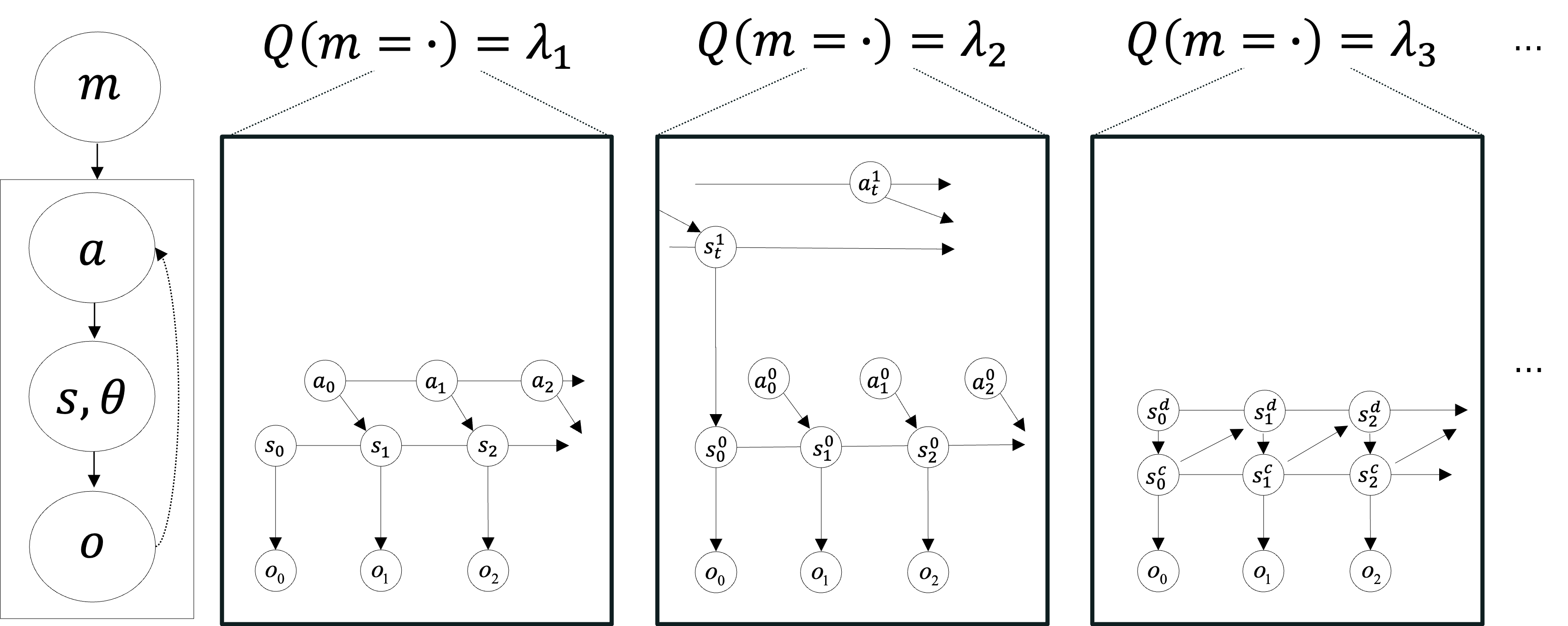}
    \caption{\textbf{The simplest generative model.} This figure describes the simplest possible generative model, where the causal network of the environment is constant over time. That is the structure $m$ is unknown but constant $m\equiv m_0$ (compare with Figure \ref{fig: complex}). This constant structure has parameters and states that yield observations. Some of those states may be controllable, i.e. active states. The ensuing generative model is summarized on the left. The structure learning problem corresponds to optimizing an approximate posterior about the unknown structure given some data $Q(m_0)\approx P(m_0 \mid d)$. The right panel shows three possible structures along with their respective probability under the approximate posterior belief.}
    \label{fig: simplest}
\end{figure}

In the simplest case, the causal structure of the environment is constant over time. In this case the simplest appropriate world model 
describes the causal network as a static hyperparameter that needs to be inferred, whence the agent only influences the states and parameters of the external process through action. We illustrate this generative model in Figure \ref{fig: simplest}.

\begin{figure}
    \centering
    \includegraphics[width=\linewidth]{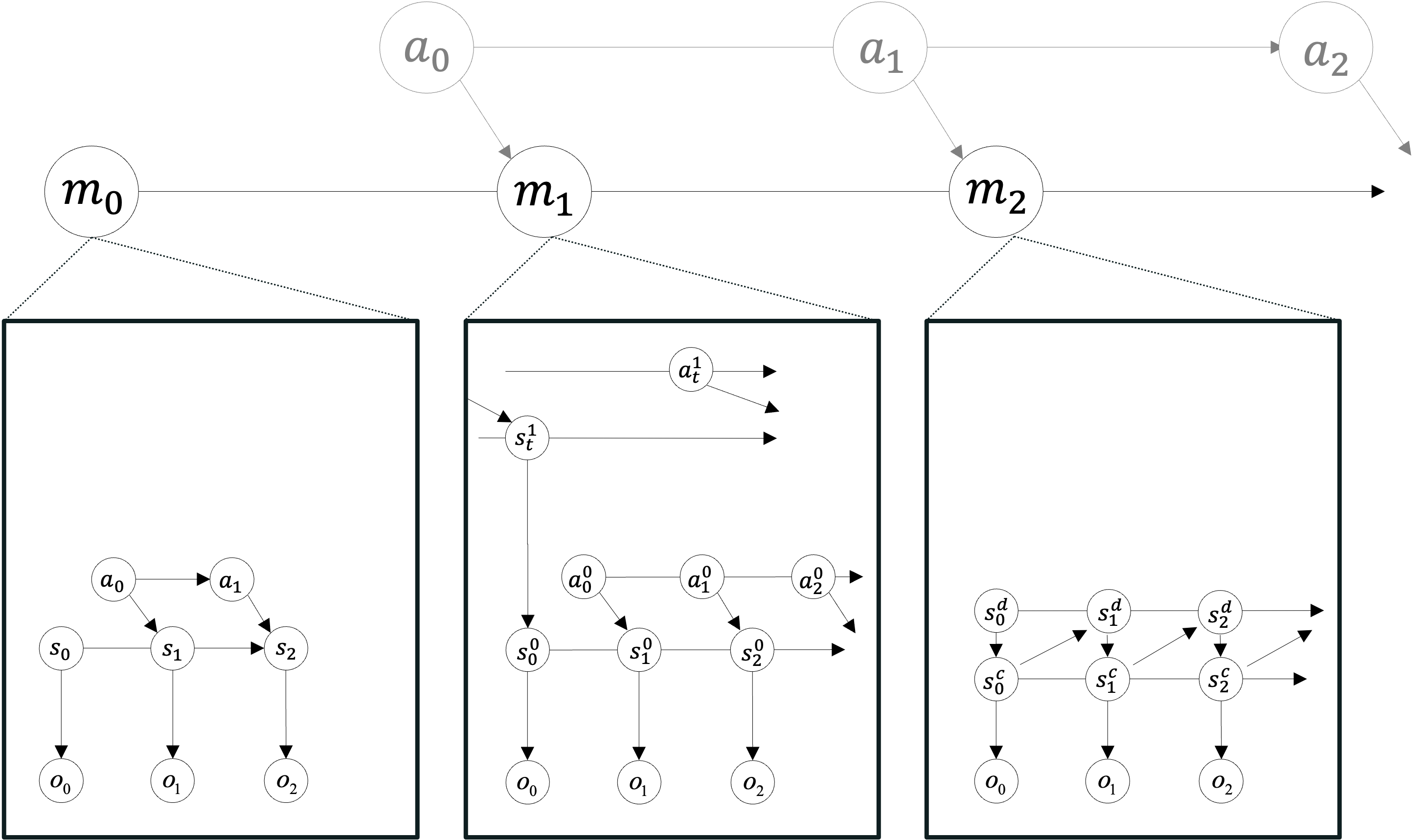}
    \caption{\textbf{More complex generative models.} These generative models describe environments where the structure or latent representations change over time; that is, when $m$ is not constant. In black (resp. black and gray) is the case where the structure is not controllable (resp. controllable): this is a hidden Markov model (resp. partially observed Markov decision process) over Bayesian networks that support efficient action, perception and learning.}
    \label{fig: complex}
\end{figure}

More generally, the causal relationships in the environment may evolve over time and may or may not be controllable by the agent. This is, for example, the case with games containing levels of increasing difficulty, where each level varies in complexity---or curriculum learning environments that gradually introduce more complex concepts as learning progresses~\cite{bengioCurriculumLearning2009,sovianyCurriculumLearningSurvey2022a}.
The causal network of the environment may be controllable, for example, when taking a particular action removes (e.g. kills) another object or agent in a game.
To represent both of these scenarios agents require more complex generative models: hidden Markov models and POMDPs on the causal network, which lead the agent to optimize beliefs about the (past, present and future) causal network of the world, which may or may not be conditioned on a course of action (i.e. during planning \eqref{eq: EFE}). Please see the illustration in Figure \ref{fig: complex}.

\subsection{Looking forward}

\paragraph{Toward a universal, interpretable, agentic class of models:}
We have described a class of models that approximates a very large class of stochastic processes on both discrete and continuous states, and which may serve as a generic model class for agent environment interactions. This class of models is very expressive while being sufficiently sparse that it can plausibly be searched \cite{fristonSupervisedStructureLearning2023}. (This is because the causal network is largely determined from the latent representations, so that 
this finesses the combinatorial explosion of having to consider all possible causal maps for a given array of latent representations). Each model in this class supports efficient action and perception, and may be interpretable.

\paragraph{The prior over models:} This class of models is infinite, so one has the choice of a non-parametric prior \cite{gershmanTutorialBayesianNonparametric2012}, or a prior on a large finite subclass. It begs the of question what kind of priors support the most efficient kind of inference, and which are most biologically plausible and apt to account for brain function? 
In any case, the prior over models should be informed by information geometric consistency constraints (i.e. local constraints) and core knowledge considerations \cite{pouncyInductiveBiasesTheorybased2022} (i.e. non-local constraints).

\paragraph{Approximate inference over models:} We surveyed various approaches to approximate inference over models in Section \ref{sec: approx inf}, concluding with a range of questions: which of these methods is most apt to operate fast and online, as would be required of structure learning agents? Which of these methods (if any) is biologically plausible, and may be most apt to model structure learning in the human brain? Are particle approximate posteriors more biologically plausible?

\paragraph{Refining the model class:}
While a promising step, the model class we discussed is likely to be insufficient for many purposes, and future work should test its limitations, actively seeking to make it more expressive, while keeping it sufficiently coarse for efficient structural inference. 
One interesting additional constraint to adequately reduce this model class is imposing a scale-free aspect to these hierarchical structures, which can be motivated by appealing to the renormalisation group and by biomimetic considerations \cite{fristonPixelsPlanningScalefree2024}. The resulting (reduced) class of structures is still expressive enough to model video from raw pixels and sound files, and plan from pixel data \cite{fristonPixelsPlanningScalefree2024}. Core knowledge should further help in refining the model building blocks. For example, noting that agents can interact at a distance while objects cannot, so that two objects in the external world lead to more causal independencies than two agents; formalizing objects and agents in this way will allow us to consider animate vs inanimate latent factors that come with distinctive causal independencies that need not be relearned every time. The same goes for all other systems of core knowledge \cite{spelkeCoreKnowledge2007,spelkeWhatBabiesKnow2022}.

\paragraph{Amortization with deep neural networks:} Despite the absence of traditional neural networks in the hierarchical models described here, deep neural networks can be important for amortizing certain inferences over states, parameters and structure \cite{mazzagliaFreeEnergyPrinciple2022}. This is particularly fitting if we consider thinking fast and slow---that is, Type I and Type II reasoning à la Kahneman \cite{kahnemanThinkingFastSlow2013}---as being instantiated by amortized and iterative inference, respectively \cite{tscshantzHybridPredictiveCoding2023}.

\section{AI alignment}
\label{sec: alignment}

We now shift gears and discuss AI alignment as a potential application of structure learning active inference agents. AI alignment refers to the challenge of ensuring that artificial intelligence systems behave in ways that are aligned with human values and intentions. This problem has gained increasing attention as AI systems become more capable and autonomous, with potential far-reaching consequences for humanity. AI safety, a closely related topic, focuses on developing AI systems that are robust, reliable, and safe in their operations. These topics have been extensively explored in seminal works such as \textit{Superintelligence} by Nick Bostrom~\cite{bostrom2014superintelligence} and \textit{Human Compatible} by Stuart Russell~\cite{russell2019human}, which highlight the potential risks and challenges of advanced AI systems.

In what follows, we approach AI alignment through the lens of active inference and structure learning, using Asimov's Three Laws of robotics~\cite{asimovRobot1950} as a simple illustrative example. Our focus is on providing new ways of thinking about the AI alignment problem, rather than recommending a specific solution to be implemented. The main conceptual point is that we can frame alignment as taking actions that comply with the other's preferences, and we can infer those preferences through structure learning, which, in the language of psychology, corresponds to instantiating a sophisticated form of theory of mind. What follows is necessarily conceptual, as a practical implementation would first require structure learning active inference agents as presented in Section \ref{sec: SLA} that have the capability to represent other agents with different generative models---a capability that remains to be developed.

\subsection{Well-being, alignment and cautious AI}
\label{item: FEP alignment}

For building safe and aligned AI systems active inference offers three conceptual deliverables:

\paragraph{Defining well-being and harm:} 
In active inference, the well-being of an agent is quantified by the (log) evidence for its generative model of the world (in practice the bound on log evidence), and harm is quantified by its negative.
At time $t$:
\begin{equation}
\label{eq: harm}
    \operatorname{well-being}(t)\triangleq\log P(d)\triangleq -\operatorname{harm}(t),
\end{equation}
where $d\triangleq d(t)$ is the data entertained by the agent at time $t$; necessarily a subset of past and present states of the agent (boundary and internal states). Note that well-being, when expressed in this way, is quantified in terms of natural units of information (nats). This definition of well-being is fairly established in the active inference literature \cite{kiversteinPlayfulnessMeaningfulLife2023,waughResilienceAbilityMaintain2023,smithComputationalNeurosciencePerspective2022,millerPredictiveDynamicsHappiness2022}. Additionally, when harm is quantified in this way, the resulting equations of motion from active inference can reproduce well-known empirical phenomena observed in biological collectives under harm \cite{ueltzhofferVariationalFreeEnergy2021}, lending additional (post-hoc) validity to this definition. 

\paragraph{Alignment:} On this view, \textit{being aligned with another is just having a high model evidence under the other's model of the world}. This means conforming with the other's model of the world, which, in active inference describes how things should ideally behave from the other's perspective (recall Section \ref{sec: aif}). For instance, an AI assistant that accurately completes tasks as desired and intended would have high model evidence under a human's generative model of helpful behavior. Conversely, an AI system that acts in unexpected or harmful ways would be highly surprising under this model and thus misaligned. This perspective on alignment emphasizes the importance of learning and respecting the preferences and expectations embedded in others' world models, which is a key challenge in developing safe and beneficial AI.
    
\paragraph{Cautious actions:} The expected free energy objective \eqref{eq: EFE} for action selection in active inference promotes cautious behavior. The risk component of the expected free energy acts as a mode-seeking objective (as a reverse KL divergence \cite{minkaDivergenceMeasuresMessage2005}), causing the agent to avoid low-probability regions under its preferred distribution. Additionally, agents minimize ambiguity by exploring to gain observations that reveal the external process, thereby improving the accuracy of risk quantification (i.e., assessment). In novel environments, ambiguity reduction initially dominates, driving exploratory behavior until the agent has sufficient information to effectively minimize risk \cite{fristonActiveInferenceProcess2017}. For example, an AI assistant might initially ask clarifying questions about a user's request, ensuring it understands the task correctly and avoids potential misinterpretations. This balance between exploration and exploitation allows for adaptive, context-aware decision-making that aligns with the agent's learned preferences and environmental understanding.

Beyond these conceptual points, active inference also offers algorithmic advances for building safer and more robust AI systems: e.g.~\cite{dacostaHowActiveInference2022a,lanillosActiveInferenceRobotics2021}.

\subsection{Toward empathetic AI}
\label{sec: empathetic ai}

We now define an \textit{empathetic agent} as an agent that not only models the objects in its external world, but also the other agents within it as well as their mental (i.e. internal) states. This effectively recognizes that the external world comprises other agents' actions, observations and internal states, which may be inferred from data along with all other external states in the shared environment. Technically, an agent that has beliefs about another's mental state is capable of empathy. In the language of cognitive psychology, empathetic AIs have \textit{theory of mind} \cite{frith_theory_2005}. 

\begin{figure}
    \centering
    \includegraphics[width=0.8\linewidth]{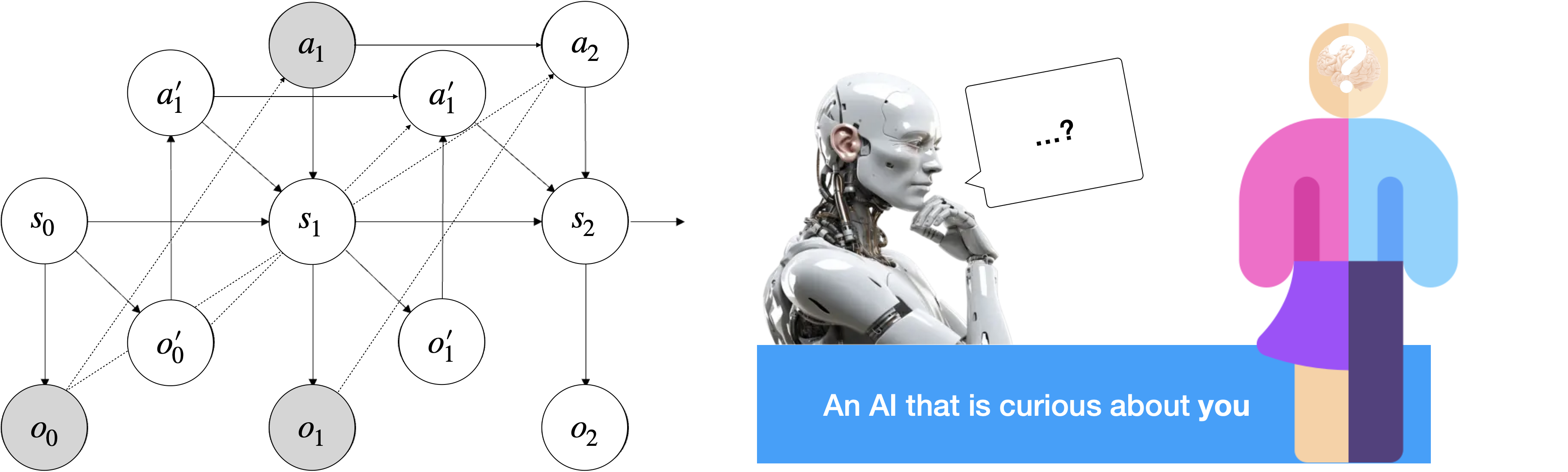}
    \caption{\textbf{Empathetic AIs.} \textit{Left:} This illustrates a simple world model of an empathetic agent in a world that comprises another agent. From the perspective of the empathetic agent, the other agent is part of the external process; however, for modeling purposes it may represent it distinctively from other inanimate external states (cf. core knowledge about agents). The other agent is formed of a sensory process $o'$, an active process $a'$ and an internal process $\mu'$ (not shown) that interact with the remaining external states as illustrated. In inferring the external process, an empathetic agent must infer the other's sensations, actions and internal states (i.e. mental state), inferences which it may in turn exploit to infer the other's world model. \textit{Right}: An empathetic agent that takes actions to minimize expected free energy would be impelled to resolve ambiguity about the other agent (i.e. \eqref{eq: EFE}). This means that an empathetic active inference agent is naturally curious about other agents. The agent's actions, such as asking questions about the other's mental state or plans, serve to reduce uncertainty and could be used for cooperative or competitive purposes depending on the agent's objectives.}
    \label{fig: empathetic AI}
\end{figure}

Instantiating this in practice is a difficult problem, and may require entertaining generative world models as in Figure \ref{fig: empathetic AI} (left panel). An empathetic agent may, in principle, infer another agent's world model from the other's inferred observations and actions (assuming, for instance, that it takes actions that minimize expected free energy---or any other objective---and solving the inverse problem). Note that assuming that the other possesses the same generative model as oneself ("you are like me") greatly simplifies the process by which one forms beliefs about another, because my model of me becomes my model of you \cite{fristonDuetOne2015}.

Learning the structure of other agents' world models becomes necessary for perspective taking, when another's generative model is structurally different from oneself, e.g. as might be necessary for an AI teacher or an AI learner. There are well-studied systems of core knowledge about how humans represent other agents and their mental states that could be leveraged for practically learning the structure of other agents' world models \cite{spelkeWhatBabiesKnow2022}. This process of inferring and learning the structure of other agents' world models is precisely a question of (Bayesian) structure learning, and will necessitate extending these concepts to the domain of social cognition \cite{ohataInvestigationSenseAgency2020} and multi-agent interactions \cite{hyland2024freeenergy}. This may sound ambitious; however, the basic procedures are now used routinely in computational psychiatry, where the generative model that best explains a patient's choice behavior is identified using Bayesian model selection. This is known as computational phenotyping. The proposal here is to endow artificial agents with this phenotyping ability.

We note that empathetic active inference agents are naturally curious about and cautious with other agents. Specifically, expected free energy minimizing actions must resolve ambiguity about other agents, while being risk-averse with respect to achieving their goals (i.e. \eqref{eq: EFE}). Under these imperatives, a preferred course of action might be seeking to communicate with other agents, to resolve uncertainty about their states, world model, well-being and future plans, etc---see Figure \ref{fig: empathetic AI} (right panel). 

Crucially for alignment, \textit{an empathetic AI that forms beliefs about others' world models would then be able to infer others' level of well-being or harm \eqref{eq: harm}}. Of course, the estimate of another’s well-being will depend on the agent’s model of the other, and this model (e.g. the coarse-graining of its representations) needs to be carefully optimized. Luckily, this optimization is precisely the optimization of model evidence \eqref{eq: accuracy complexity} that underlies Bayesian structure learning, in the sense of finding the maximally accurate and minimally complex representation of the other in relation to available data.



However, it is crucial to recognize that empathy \textit{alone} does not guarantee benevolence or safety. Competitive or even adversarial agents may benefit from sophisticated models of others for instrumental reasons. In particular they may use these to manipulate, or deceive others to achieve their goals~\cite{ashton2022problem,franklin2022recognising}. The key might be to prescribe agent's actions to be benevolent in the sense of complying with the other's preferences. However, note that benevolent agents may still choose to deceive, as it may happen to be beneficial to manipulate humans in the process of optimizing their well-being (cf. white lies). 

\subsection{The Three Laws of Robotics}\label{sec:3laws}

Being aligned with another is just conforming to the other's model of the world. But a rock at rest being aligned with me does not mean the rock is being helpful. We now explore more specific approaches to AI alignment and revisit a well-known concept from science fiction: Asimov's Three Laws of Robotics \cite{asimovRobot1950}. These laws provide an interesting and illustrative thought experiment for our discussion of empathetic agents. The \textit{Three Laws of Robotics} are:
\begin{quote}
\begin{enumerate}
    \item A robot may not injure a human being or, through inaction, allow a human being to come to harm. 
    \item A robot must obey orders given it by human beings except where such orders would conflict with the First Law.
    \item A robot must protect its own existence as long as such protection does not conflict with the First or Second Law.
\end{enumerate}
    - Asimov (1950) \textit{I, Robot} \cite{asimovRobot1950}.
\end{quote}
While these laws are not a practical solution to AI alignment in an of themselves,\footnote{Indeed, Asimov conceived and employed these laws as a narrative device to explore the complexities and potential failures of simple ethical rules for AI~\cite{asimovRobot1950}, as is demonstrated in many of his stories featuring the Three Laws.} they offer a starting point for illustrating how we might formalize ethical constraints within an active inference framework.

Let's examine how we might approach the spirit of the First Law using the concepts discussed so far. Minimizing harm---respectively maximizing well-being in the sense of \eqref{eq: harm}---supports the First Law, 
and may turn out to be stronger, possibly already including, e.g. the Second and Third laws, as we will see later, and thus this implies courses of action beyond preventing harm in the colloquial sense.

\paragraph{The First Law of Robotics:}
To mathematically instantiate the First Law, first \textit{consider an empathetic agent whose sole preferences are to prevent another agent (the \emph{target agent}) from coming to harm}. The empathetic agent's actions could be taken to minimize the expected free energy cost function \eqref{eq: EFE}, where we discard all external variables besides the target agent's harm
\begin{equation}
\label{eq: First Law}
\G= \underbrace{\dkl[\overbrace{P(\text{FE} \mid  a,d)}^{\text{predicted}} \mid \overbrace{P(\text{FE} \mid d)}^{\text{preferred}} ]}_{\text{Risk}}+\underbrace{\mathbb E_{P(\text{FE}  \mid  a,d)}[-\log P(o \mid \text{FE},d)]}_{\text{Ambiguity}}.
\end{equation}
Here $\text{FE}$ is the harm (or negative well-being) of the target agent---the only external variable that is retained in the expected free energy to govern the empathetic agent's actions. \eqref{eq: First Law} means that the empathetic agent's only imperatives are to take actions to match a preferred distribution over the target agent's well-being and to minimize ambiguity about their well-being. 

A version of the First Law could be achieved by designing the preferred distribution over the target agent's harm in \eqref{eq: First Law} to assign greater probability to low harm than to high harm. For example, an exponentially decaying probability distribution over harm so that zero or low harm would exponentially preferred to high harm, or by (soft-)thresholding the target's well-being to desirable values. In any case, the fact that the preferred distribution in \eqref{eq: First Law} depends on the history of data $d$ means that preferences may be learned over time in a data-driven way \cite{sajidActiveInferencePreference2022,dacostaActiveInferenceModel2022}. To appropriately model the target's level of harm, the empathetic agent would have to model many other external variables; however the only imperatives driving its actions would be to resolve ambiguity---and realize the preferred distribution---about the target's well-being.

Optimizing well-being in the short term can be very different from optimizing well-being in the longer term, and scoring courses of action according to \eqref{eq: First Law} entails a planning horizon that defines the time-span over which the empathetic agent seeks to improve the target agent's well-being. This time-span is the temporal depth in the empathetic agent's generative model of harm (cf. Section \ref{sec: expressivity stochastic processes}). Because the empathetic agent bases its inferences on the target's world model, the empathetic agent's temporal horizon for well-being is its estimate for the target's own temporal horizon. This may not be the desired behavior depending on the context, as it may be desirable to protect the target agent even from events it may not be able to foresee but would predictably care about (or be harmed by) at the time they occur. However, this problem is finessed in multi-agent systems when acting to fulfill the preferences of multiple target agents, as the planning time horizon is expected to become the supremum of the target's horizons. 


\paragraph{The First Law in multi-agent settings:}
In an environment with $n$ target agents, we could formulate the First Law for multiple agents by replacing $\text{harm}$ in \eqref{eq: First Law} with $(\text{harm}_1,\ldots,\text{harm}_n)$, which represents the joint well-being of target agents. The empathetic agent would therefore have to predict each target's level of well-being as well as maintaining a joint preference distribution over these variables. One possible implementation of this is to treat each target's well-being equally and independently, leading to a factorized joint preference distribution: $P(\text{harm}_1,\ldots,\text{harm}_n \mid d)=\prod_{i=1}^{n}P(\text{harm}_i\mid d)$ where the preference distribution over each individual's harm is identical for all individuals. If we further specify that the target agents are the humans in the environment, then \eqref{eq: First Law} becomes an instantiation of Asimov's First Law of robotics: a specification of cautious actions with the intent to prevent any humans from coming to harm. However, we note that the specific choice of the joint distribution is itself a complex ethical and societal question, which lies beyond this paper's scope. 

\paragraph{The Second and Third Laws:}
If we reintroduce the external variables back into expected free energy \eqref{eq: First Law} with a flat (or maximally entropic) prior preference, the empathetic agent's actions should remain largely unchanged---in particular remain consistent with the First Law---but crucially, we may allow the preferences over these additional variables to be learned from experience. Since preferences in active inference are learned by maximizing model evidence, these additional preferences (i.e. incentives \cite{Everitt2019UnderstandingAI,everitt2021agentincentivescausalperspective}) would be learned in such a way that they are most conducive to satisfy the hard constraint of maximizing others' well-being encoded in the First Law.
\begin{itemize}
    \item \textbf{The Second Law}: From the observation that disobeyance induces harm, in the sense of \eqref{eq: harm}, a robot would learn that in order to minimize the harm to other humans, it should obey their orders unless this conflicts with the First Law.
    \item \textbf{The Third Law}: From the observation that a robot must maintain its existence to actively minimize others' harm, a robot might learn to protect its own existence as long as this does not cause harm by conflicting with the First or Second Law.
\end{itemize}
Note that in this version of the Three Laws, the Second and Third Laws emerge from, and are learned to enable the First Law. 
An important nuance is that empathetic agents would follow what would make their target agents' better off---even unbeknownst to the target agents---even if these explicitly order otherwise. Possibly, this implies that empathetic agents' interventional capabilities should be limited until they have acquired a sufficient capability to understand what is beneficial for its human targets (through theory of mind in the multi-agent context).
Beyond a human-centric approach, it may be more desirable for empathetic agents to seek to maximize the well-being of all other agents in their ecosystem. This might be achieved by modifying the First Law \eqref{eq: First Law} to consider the well-being of all agents not limited to human agents.

To be sure, we are a long way away from practically being able to instantiate these Laws beyond toy examples and examining them in detail invites a range of challenges. However we hope that expressing these and similar ideas in a more formal language will aid practical investigations.

\subsection{Alignment beyond perspective taking}

While we have primarily discussed alignment in terms of perspective-taking and theory of mind, it is important to recognize that assuming an explicit model of other agents is not necessary for alignment. For example, the biome in our gut is generally aligned with ourselves, and its bacterial components are conceivably to some extent aligned with each other, yet it seems unreasonable to believe that any of them possess a sophisticated theory of mind about other bacteria or their host.

Indeed, from the perspective of active inference, the more fundamental question for alignment is: what kind of agent interactions lead to well-being maximization for each member of a group or ecosystem? This regime, dubbed the `free-energy equilibrium' \cite{hyland2024freeenergy}, generalizes the classical game-theoretic notion of Nash equilibrium to boundedly-rational agents. 
These equilibria are starting to be formally studied \cite{hylandFreeEnergyEquilibriaTheory2024}, and understanding these opens the possibility of engineering agents that would bring their ecosystem closer to a free energy equilibrium. 

Numerical studies suggest that populations may naturally converge to free energy equilibria, where all agents share the same generative model (and preferences), albeit possibly at an evolutionary timescale \cite{isomuraBayesianFilteringMultiple2019,fristonFederatedInferenceBelief2024}. In this case, each agent's goal is the group's (shared) goal and empathy is an emergent property \cite{fristonDuetOne2015}. Clearly, these numerical studies need to be extended to empathetic agents and agents that learn their own preferences \cite{sajidActiveInferencePreference2022,dacostaActiveInferenceModel2022}.

In summary, while empathetic agents may be one way of achieving alignment, other routes are possible. Free energy equilibria provide a framework for alignment in systems where explicit perspective-taking may not be feasible or necessary. This could be particularly valuable in many-agent systems or in scenarios where agents' cognitive capabilities vary widely. Future work could explore how free energy equilibria relate to other concepts in AI alignment, and how they might be applied to practically designing aligned AI systems.

\subsection{Related work}

Related work uses causal models of agent-environment interactions, but with the aim of analysing agent's incentives~\cite{everitt2021agentincentivescausalperspective,Everitt2019UnderstandingAI}. This provides a complementary perspective on analyzing and designing AI systems with desirable incentive structures, and aligns well with our discussion of theory of mind, including its potential for benevolent as well as adversarial uses. The structure learning approach we develop may provide a way to dynamically construct these causal networks and go beyond identifying the mere presence or absence of various properties such as incentives~\cite{everitt2021agentincentivescausalperspective}, intentions~\cite{ward2024reasons}, and deception~\cite{ward2024honesty}, by \textit{quantifying} these phenomena.

The challenge of avoiding unintended consequences of action, e.g. \cite{Krakovna2018PenalizingSE}, also resonates with the desirability of risk-averse agents. The concept of penalizing actions that lead to significant, irreversible and potentially harmful changes is in line with the risk-averse behavior produced by expected free energy minimization in active inference.

Our work builds upon Bayesian models of theory of mind. Classical work demonstrates Bayesian inference of others' mental states and reward functions \cite{baker_bayesian_2011}, and that empathy can be straightforward when another has the same generative model as oneself \cite{fristonDuetOne2015}. Recent work explores multi-agent cooperation through utility function inference \cite{kleiman-weiner_evolving_2025}, but maintain only a fixed repertoire of agent profiles for classification. In all cases, approaches assume known generative model structures, whereas we emphasize structure learning about agents whose generative model is unknown. Furthermore, in contrast to previous work, we emphasise the importance of epistemic actions (e.g. asking questions) to gain information about other agents' generative model and preferences.

\section{Discussion}
\label{sec: discussion}

In this paper, we aimed to lay out a map of research questions to be addressed for scaling a naturalistic approach to aligned artificial intelligence, along with paths forward. We aimed to be inclusive of all naturalistic approaches to intelligence, and in doing so, made a particular commitment to a first principles approach known as active inference. This framework offers a coherent approach to the varied problems of structure learning and alignment. Here, we take a step back and discuss the commitments underlying each section when considered on its own:

\paragraph{A first principles approach:} Active inference can be arrived at from two opposite directions: the bottom-up, inductive and historical approach, which is where theories of brain function were continuously refined and generalized to account for various empirical phenomena \cite{raoPredictiveCodingVisual1999,knillBayesianBrainRole2004,fristonFreeenergyBrain2007,fristonFreeenergyPrincipleUnified2010}. A top-down, deductive approach, through the nascent field of physics called Bayesian mechanics \cite{ramsteadBayesianMechanicsPhysics2022,dacostaBayesianMechanicsStationary2021a,fristonFreeEnergyPrinciple2023a}, which bridges elementary descriptions of particles, things and agents in the physical world, with descriptions of inference. From a theoretical perspective, there are many opportunities to further develop this top-down approach, e.g. 1) by characterizing the subclasses of physical systems that exhibit higher-order cognitive phenomena \cite{fristonPathIntegralsParticular2023,dacostaBayesianMechanicsMetacognitive2024,sandved-smithMetacognitiveParticlesMental2024}, and 2) by making these formulations mathematically rigorous, using purposefully developed tools in the theory of stochastic processes \cite{dacostaTheoryGeneralisedCoordinates2024}.

\paragraph{Bayesian structure learning agents:} We adopted the Bayesian perspective on structure learning that follows from active inference. Namely, the external world, which is the data generating process, is a stochastic process, which may be summarized as a set of random variables and their causal relationships. Because these random variables and causal relationships are unknown, they must be inferred from data. Approximate Bayesian inference, however, is not the end but a means toward the end of optimizing the evidence for a generative model of the data-generating process, which technically corresponds to a minimal-length description---i.e. compression---of the data. When considering the agentic setting where new data continuously arrives and the agent acts upon the world, we assumed that planning was done using the world model by combining a mixture of exploration and exploitation using information-theoretic objectives \cite{hafnerActionPerceptionDivergence2020}.

\paragraph{AI safety and alignment:} Finally, the last section makes heavy use of the active inference framework by exploiting the fact that in active inference, the world model of the agent supplies the agent's preferences (i.e. expectations are preferences), so that simply learning this world model tells one how to be aligned with another. Another important feature here is the expected free energy objective for selecting actions, which prescribes cautious behavior for the agent, supporting safety and alignment.

\paragraph{Convergence in computational cognitive science \& AI:} In summary, this perspective can be seen to rest on three pillars, which are facets of a same underlying phenomenon: evidence maximization for a world model, model-based planning combining exploration and exploitation, and approximate Bayesian inference about the external environment. These are common commitments in cognitive science and AI, and particularly resonate with the perspective of several AI experts seeking next generation systems through a paradigm shift \cite{scholkopfCausalRepresentationLearning2021,lecunPathAutonomousMachine2022}.

\section{Conclusion}

We have exposed principles toward more scalable aligned AI agents that come to represent their worlds, which may guide AI research. 
These principles leverage converging approaches to understanding intelligence, synthesizing ideas across mathematics, physics, statistics and cognitive science.
We have framed much of the foregoing narrative using active inference, a first principles approach to describing natural intelligence. Practically, this implies a commitment to maximizing the evidence for a generative world model, model-based planning combining exploration and exploitation, and approximate Bayesian inference about the external world and its causal structure---three common commitments in computational cognitive science and AI.
Attending this perspective are numerical studies building more scalable, capable and aligned systems based on the ideas discussed herein.

\subsection*{Acknowledgements}

LD thanks Yoshua Bengio and Bernhard Schölkopf for hosting research stays where some of this work was conducted. The authors are indebted to many for interesting discussions which helped shape this manuscript: in particular to Yoshua Bengio, Bernhard Schölkopf, Tristan Deleu, Alex Hernandez-Garcia, Moksh Jain, Oussama Boussif, Léna Néhale-Ezzine, Xiaoyin Chen, Andrea Dittadi, Guillaume Dumas, Zahra Sheikhbahaee, George Deane, Daphne Demekas, Lars Sandved-Smith, Tommie Tosato, Jonas Mago, Mahault Albarracin, Eric Elmoznino, Samuel Gershman, Joshua B. Tenenbaum, Pedro Tsividis, Alexander Tschantz, Tim Verbelen, Tommaso Salvatori and Conor Heins.

\subsection*{Funding information}
A.R. is funded by the Australian Research Council (Ref: DP200100757). A.R. is also funded by Australian National Health and Medical Research Council Investigator Grant (Ref: 1194910). A.R. is affiliated with The Wellcome Centre for Human Neuroimaging supported by core funding from Wellcome [203147/Z/16/Z]. A.R. is a CIFAR Azrieli Global Scholar in the Brain, Mind \& Consciousness Programme.

\subsection*{Competing interests}

The authors declare no competing interests.

\appendix

\section{Bayesian model reduction: practical summary}
\label{app: BMR}

Given some data $d$, one selects a base prior $P_\lambda(\eta)$ (usually maximally entropic in the family of priors, see below why). One then computes a corresponding approximate posterior $Q_{\lambda}(\eta \mid d)\approx P_{\lambda}(\eta \mid d)$. From \eqref{eq: BMR},
\begin{equation}
\label{eq: practical BMR}
    \frac{ P_{\lambda'}(d)}{ P_{\lambda}(d)}\approx \mathbb{E}_{Q_{\lambda}(\eta \mid d)}\left[\frac{P_{\lambda'}(\eta)}{P_{\lambda}(\eta)}\right].
\end{equation}
For several distribution classes, the expectation in \eqref{eq: practical BMR} has a closed form so that it is straightforward to maximize it with respect to $\lambda'$ \cite{fristonBayesianModelReduction2019}.
When a closed form is unavailable, one samples from the approximate posterior $Q_{\lambda}(\eta \mid d)$---as is done easily with GFlowNets \cite{bengioFlowNetworkBased2021,malkinGFlowNetsVariationalInference2022} or particle approximate posteriors \cite{saeediVariationalParticleApproximations2017}---to obtain a Monte-Carlo estimator of the expectation for any given $\lambda'$. In either case, the goal is to maximize the ratio of evidences with respect to $\lambda'$, resulting a new prior and improved generative model $P(d\mid \eta)P_{\lambda'}(\eta)$ that can be used for treating the next incoming batch of data. 

\paragraph{Why is this called model \textit{reduction}?} 
If the base prior is not maximally entropic $P_\lambda(\eta)$, then this will also be the case for the approximate posterior, but then the Monte-Carlo estimator of the expectation in \eqref{eq: practical BMR} has a higher variance, which makes the method less practical. Hence, the base prior is typically chosen to be maximally entropic so that applying this method reduces prior and model entropy.

\bibliographystyle{unsrt}
\bibliography{bib,bib2}

\begin{thebibliography}{100}

\bibitem{dacostaActiveInferenceDiscrete2020}
Lancelot Da~Costa, Thomas Parr, Noor Sajid, Sebastijan Veselic, Victorita Neacsu, and Karl Friston.
\newblock Active inference on discrete state-spaces: {{A}} synthesis.
\newblock {\em Journal of Mathematical Psychology}, 99:102447, December 2020.

\bibitem{parrActiveInferenceFree2022}
Thomas Parr, Giovanni Pezzulo, and Karl~J. Friston.
\newblock {\em Active {{Inference}}: {{The Free Energy Principle}} in {{Mind}}, {{Brain}}, and {{Behavior}}}.
\newblock MIT Press, Cambridge, MA, USA, March 2022.

\bibitem{buckleyFreeEnergyPrinciple2017}
Christopher~L. Buckley, Chang~Sub Kim, Simon McGregor, and Anil~K. Seth.
\newblock The free energy principle for action and perception: {{A}} mathematical review.
\newblock {\em Journal of Mathematical Psychology}, 81:55--79, December 2017.

\bibitem{helmholtzHelmholtzsTreatisePhysiological1962}
Hermann von Helmholtz and James P.~C Southall.
\newblock {\em Helmholtz's Treatise on Physiological Optics.}
\newblock Dover Publications, New York, 1962.

\bibitem{raoPredictiveCodingVisual1999}
Rajesh P.~N. Rao and Dana~H. Ballard.
\newblock Predictive coding in the visual cortex: A functional interpretation of some extra-classical receptive-field effects.
\newblock {\em Nature Neuroscience}, 2(1):79--87, January 1999.

\bibitem{knillBayesianBrainRole2004}
David~C. Knill and Alexandre Pouget.
\newblock The {{Bayesian}} brain: The role of uncertainty in neural coding and computation.
\newblock {\em Trends in Neurosciences}, 27(12):712--719, December 2004.

\bibitem{fristonTheoryCorticalResponses2005}
Karl Friston.
\newblock A theory of cortical responses.
\newblock {\em Philosophical Transactions of the Royal Society B: Biological Sciences}, 360(1456):815--836, April 2005.

\bibitem{fristonFreeenergyBrain2007}
Karl~J. Friston and Klaas~E. Stephan.
\newblock Free-energy and the brain.
\newblock {\em Synthese}, 159(3):417--458, November 2007.

\bibitem{fristonFreeenergyPrincipleUnified2010}
Karl Friston.
\newblock The free-energy principle: A unified brain theory?
\newblock {\em Nature Reviews Neuroscience}, 11(2):127--138, February 2010.

\bibitem{fristonFreeEnergyPrinciple2023a}
Karl Friston, Lancelot Da~Costa, Noor Sajid, Conor Heins, Kai Ueltzh{\"o}ffer, Grigorios~A. Pavliotis, and Thomas Parr.
\newblock The free energy principle made simpler but not too simple.
\newblock {\em Physics Reports}, 1024:1--29, June 2023.

\bibitem{parrComputationalNeurologyMovement2021a}
Thomas Parr, Jakub Limanowski, Vishal Rawji, and Karl Friston.
\newblock The computational neurology of movement under active inference.
\newblock {\em Brain}, 144(6):1799--1818, June 2021.

\bibitem{adamsComputationalAnatomyPsychosis2013}
Rick~A. Adams, Klaas~Enno Stephan, Harriet~R. Brown, Christopher~D. Frith, and Karl~J. Friston.
\newblock The {{Computational Anatomy}} of {{Psychosis}}.
\newblock {\em Frontiers in Psychiatry}, 4, 2013.

\bibitem{fristonActiveInferenceProcess2017}
Karl Friston, Thomas FitzGerald, Francesco Rigoli, Philipp Schwartenbeck, and Giovanni Pezzulo.
\newblock Active {{Inference}}: {{A Process Theory}}.
\newblock {\em Neural Computation}, 29(1):1--49, January 2017.

\bibitem{fristonGraphicalBrainBelief2017}
Karl~J. Friston, Thomas Parr, and Bert {de Vries}.
\newblock The graphical brain: {{Belief}} propagation and active inference.
\newblock {\em Network Neuroscience}, 1(4):381--414, December 2017.

\bibitem{isomuraCanonicalNeuralNetworks2022}
Takuya Isomura, Hideaki Shimazaki, and Karl~J. Friston.
\newblock Canonical neural networks perform active inference.
\newblock {\em Communications Biology}, 5(1):1--15, January 2022.

\bibitem{isomuraExperimentalValidationFreeenergy2023}
Takuya Isomura, Kiyoshi Kotani, Yasuhiko Jimbo, and Karl~J. Friston.
\newblock Experimental validation of the free-energy principle with in vitro neural networks.
\newblock {\em Nature Communications}, 14(1):4547, August 2023.

\bibitem{smithStepbystepTutorialActive2022}
Ryan Smith, Karl~J. Friston, and Christopher~J. Whyte.
\newblock A step-by-step tutorial on active inference and its application to empirical data.
\newblock {\em Journal of Mathematical Psychology}, 107:102632, April 2022.

\bibitem{lanillosActiveInferenceRobotics2021}
Pablo Lanillos, Cristian Meo, Corrado Pezzato, Ajith~Anil Meera, Mohamed Baioumy, Wataru Ohata, Alexander Tschantz, Beren Millidge, Martijn Wisse, Christopher~L. Buckley, and Jun Tani.
\newblock Active {{Inference}} in {{Robotics}} and {{Artificial Agents}}: {{Survey}} and {{Challenges}}.
\newblock {\em arXiv:2112.01871 [cs]}, December 2021.

\bibitem{dacostaHowActiveInference2022a}
Lancelot Da~Costa, Pablo Lanillos, Noor Sajid, Karl Friston, and Shujhat Khan.
\newblock How {{Active Inference Could Help Revolutionise Robotics}}.
\newblock {\em Entropy}, 24(3):361, March 2022.

\bibitem{mazzagliaFreeEnergyPrinciple2022}
Pietro Mazzaglia, Tim Verbelen, Ozan {\c C}atal, and Bart Dhoedt.
\newblock The {{Free Energy Principle}} for {{Perception}} and {{Action}}: {{A Deep Learning Perspective}}.
\newblock {\em Entropy}, 24(2):301, February 2022.

\bibitem{heins2025axiomlearningplaygames}
Conor Heins, Toon~Van de~Maele, Alexander Tschantz, Hampus Linander, Dimitrije Markovic, Tommaso Salvatori, Corrado Pezzato, Ozan Catal, Ran Wei, Magnus Koudahl, Marco Perin, Karl Friston, Tim Verbelen, and Christopher Buckley.
\newblock Axiom: Learning to play games in minutes with expanding object-centric models, 2025.

\bibitem{fristonPixelsPlanningScalefree2024a}
Karl Friston, Conor Heins, Tim Verbelen, Lancelot Da~Costa, Tommaso Salvatori, Dimitrije Markovic, Alexander Tschantz, Magnus Koudahl, Christopher Buckley, and Thomas Parr.
\newblock From pixels to planning: Scale-free active inference, July 2024.

\bibitem{detinguyExploringLearningStructure2024}
Daria {de Tinguy}, Tim Verbelen, and Bart Dhoedt.
\newblock Exploring and {{Learning Structure}}: {{Active Inference Approach}} in {{Navigational Agents}}, August 2024.

\bibitem{pouncyWhatModelModelBased2021}
Thomas Pouncy, Pedro Tsividis, and Samuel~J. Gershman.
\newblock What {{Is}} the {{Model}} in {{Model-Based Planning}}?
\newblock {\em Cognitive Science}, 45(1):e12928, January 2021.

\bibitem{scholkopfCausalRepresentationLearning2021}
Bernhard Sch{\"o}lkopf, Francesco Locatello, Stefan Bauer, Nan~Rosemary Ke, Nal Kalchbrenner, Anirudh Goyal, and Yoshua Bengio.
\newblock Toward {{Causal Representation Learning}}.
\newblock {\em Proceedings of the IEEE}, 109(5):612--634, May 2021.

\bibitem{petersElementsCausalInference2017}
Jonas Peters, Dominik Janzing, and Bernhard Sch{\"o}lkopf.
\newblock {\em Elements of {{Causal Inference}}: {{Foundations}} and {{Learning Algorithms}}}.
\newblock The MIT Press, 2017.

\bibitem{gershmanLearningLatentStructure2010}
Samuel~J. Gershman and Yael Niv.
\newblock Learning latent structure: Carving nature at its joints.
\newblock {\em Current Opinion in Neurobiology}, 20(2):251--256, April 2010.

\bibitem{tsividisHumanLevelReinforcementLearning2021}
Pedro~A. Tsividis, Joao Loula, Jake Burga, Nathan Foss, Andres Campero, Thomas Pouncy, Samuel~J. Gershman, and Joshua~B. Tenenbaum.
\newblock Human-{{Level Reinforcement Learning}} through {{Theory-Based Modeling}}, {{Exploration}}, and {{Planning}}.
\newblock {\em arXiv:2107.12544 [cs]}, July 2021.

\bibitem{pouncyInductiveBiasesTheorybased2022}
Thomas Pouncy and Samuel~J. Gershman.
\newblock Inductive biases in theory-based reinforcement learning.
\newblock {\em Cognitive Psychology}, 138:101509, November 2022.

\bibitem{bleiVariationalInferenceReview2017}
David~M. Blei, Alp Kucukelbir, and Jon~D. McAuliffe.
\newblock Variational {{Inference}}: {{A Review}} for {{Statisticians}}.
\newblock {\em Journal of the American Statistical Association}, 112(518):859--877, April 2017.

\bibitem{bishopPatternRecognitionMachine2006}
Christopher~M. Bishop.
\newblock {\em Pattern Recognition and Machine Learning}.
\newblock Information Science and Statistics. Springer, New York, 2006.

\bibitem{fristonPathIntegralsParticular2023}
Karl Friston, Lancelot Da~Costa, Dalton A.~R. Sakthivadivel, Conor Heins, Grigorios~A. Pavliotis, Maxwell Ramstead, and Thomas Parr.
\newblock Path integrals, particular kinds, and strange things.
\newblock {\em Physics of Life Reviews}, August 2023.

\bibitem{dacostaBayesianMechanicsStationary2021a}
Lancelot Da~Costa, Karl Friston, Conor Heins, and Grigorios~A. Pavliotis.
\newblock Bayesian mechanics for stationary processes.
\newblock {\em Proceedings of the Royal Society A: Mathematical, Physical and Engineering Sciences}, 477(2256):20210518, December 2021.

\bibitem{dacostaTheoryGeneralisedCoordinates2024}
Lancelot Da~Costa, Natha{\"e}l Da~Costa, Conor Heins, Johan Medrano, Grigorios~A. Pavliotis, Thomas Parr, Ajith~Anil Meera, and Karl Friston.
\newblock A theory of generalised coordinates for stochastic differential equations, September 2024.

\bibitem{fristonStochasticChaosMarkov2021}
Karl Friston, Conor Heins, Kai Ueltzh{\"o}ffer, Lancelot Da~Costa, and Thomas Parr.
\newblock Stochastic {{Chaos}} and {{Markov Blankets}}.
\newblock {\em Entropy}, 23(9):1220, September 2021.

\bibitem{heinsSparseCouplingMarkov2022}
Conor Heins and Lancelot Da~Costa.
\newblock Sparse coupling and {{Markov}} blankets: {{A}} comment on "{{How}} particular is the physics of the {{Free Energy Principle}}?" by {{Aguilera}}, {{Millidge}}, {{Tschantz}} and {{Buckley}}.
\newblock {\em arXiv:2205.10190 [cond-mat, physics:nlin]}, May 2022.

\bibitem{barpGeometricMethodsSampling2022a}
Alessandro Barp, Lancelot Da~Costa, Guilherme Fran{\c c}a, Karl Friston, Mark Girolami, Michael~I. Jordan, and Grigorios~A. Pavliotis.
\newblock Geometric {{Methods}} for {{Sampling}}, {{Optimisation}}, {{Inference}} and {{Adaptive Agents}}.
\newblock volume~46, pages 21--78. 2022.

\bibitem{dacostaActiveInferenceModel2022}
Lancelot Da~Costa, Samuel Tenka, Dominic Zhao, and Noor Sajid.
\newblock Active {{Inference}} as a {{Model}} of {{Agency}}.
\newblock In {\em Workshop on {{RL}} as a Model of Agency}, 2022.

\bibitem{sajidActiveInferenceBayesian2022}
Noor Sajid, Lancelot Da~Costa, Thomas Parr, and Karl Friston.
\newblock Active inference, {{Bayesian}} optimal design, and expected utility.
\newblock In {\em The {{Drive}} for {{Knowledge}}: {{The Science}} of {{Human Information Seeking}}}. 2022.

\bibitem{fristonSophisticatedInference2021}
Karl Friston, Lancelot Da~Costa, Danijar Hafner, Casper Hesp, and Thomas Parr.
\newblock Sophisticated {{Inference}}.
\newblock {\em Neural Computation}, 33(3):713--763, February 2021.

\bibitem{fristonActionBehaviorFreeenergy2010}
Karl~J. Friston, Jean Daunizeau, James Kilner, and Stefan~J. Kiebel.
\newblock Action and behavior: A free-energy formulation.
\newblock {\em Biological Cybernetics}, 102(3):227--260, March 2010.

\bibitem{sajidActiveInferencePreference2022}
Noor Sajid, Panagiotis Tigas, and Karl Friston.
\newblock Active inference, preference learning and adaptive behaviour.
\newblock {\em IOP Conference Series: Materials Science and Engineering}, 1261(1):012020, October 2022.

\bibitem{dacostaRewardMaximizationDiscrete2023}
Lancelot Da~Costa, Noor Sajid, Thomas Parr, Karl Friston, and Ryan Smith.
\newblock Reward {{Maximization Through Discrete Active Inference}}.
\newblock {\em Neural Computation}, 35(5):807--852, April 2023.

\bibitem{weiValueInformationReward2024}
Ran Wei.
\newblock Value of {{Information}} and {{Reward Specification}} in {{Active Inference}} and {{POMDPs}}, August 2024.

\bibitem{abelThreeDogmasReinforcement2024}
David Abel, Mark~K. Ho, and Anna Harutyunyan.
\newblock Three {{Dogmas}} of {{Reinforcement Learning}}.
\newblock In {\em First {{Reinforcement Learning Conference}}}. arXiv, July 2024.

\bibitem{tenenbaumHowGrowMind2011}
Joshua~B. Tenenbaum, Charles Kemp, Thomas~L. Griffiths, and Noah~D. Goodman.
\newblock How to {{Grow}} a {{Mind}}: {{Statistics}}, {{Structure}}, and {{Abstraction}}.
\newblock {\em Science}, 331(6022):1279--1285, March 2011.

\bibitem{ullmanBayesianModelsConceptual2020}
Tomer~D. Ullman and Joshua~B. Tenenbaum.
\newblock Bayesian {{Models}} of {{Conceptual Development}}: {{Learning}} as {{Building Models}} of the {{World}}.
\newblock {\em Annual Review of Developmental Psychology}, 2(1):533--558, 2020.

\bibitem{turingCOMPUTINGMACHINERYINTELLIGENCE1950}
A.~M. TURING.
\newblock I.---{{COMPUTING MACHINERY AND INTELLIGENCE}}.
\newblock {\em Mind}, LIX(236):433--460, October 1950.

\bibitem{pearlCausality2009}
Judea Pearl.
\newblock {\em Causality}.
\newblock Cambridge University Press, Cambridge, U.K. ; New York, 2nd edition edition, September 2009.

\bibitem{federMaximumEntropySpecial1986}
M.~Feder.
\newblock Maximum entropy as a special case of the minimum description length criterion ({{Corresp}}.).
\newblock {\em IEEE Transactions on Information Theory}, 32(6):847--849, November 1986.

\bibitem{grunwaldMinimumDescriptionLength2007}
Peter D.~D. Grunwald.
\newblock {\em The {{Minimum Description Length Principle}}}.
\newblock MIT Press, March 2007.

\bibitem{landauerIrreversibilityHeatGeneration1961}
R.~Landauer.
\newblock Irreversibility and {{Heat Generation}} in the {{Computing Process}}.
\newblock {\em IBM Journal of Research and Development}, 5(3):183--191, July 1961.

\bibitem{rebaneRecoveryCausalPolytrees1987}
George Rebane and Judea Pearl.
\newblock The recovery of causal poly-trees from statistical data.
\newblock In {\em Proceedings of the {{Third Conference}} on {{Uncertainty}} in {{Artificial Intelligence}}}, {{UAI}}'87, pages 222--228, Arlington, Virginia, USA, July 1987. AUAI Press.

\bibitem{chickeringLearningBayesianNetworks1996}
David~Maxwell Chickering.
\newblock Learning {{Bayesian Networks}} is {{NP-Complete}}.
\newblock In Doug Fisher and Hans-J. Lenz, editors, {\em Learning from {{Data}}: {{Artificial Intelligence}} and {{Statistics V}}}, pages 121--130. Springer, New York, NY, 1996.

\bibitem{chickeringLargeSampleLearningBayesian2004}
David~Maxwell Chickering, David Heckerman, and Christopher Meek.
\newblock Large-{{Sample Learning}} of {{Bayesian Networks}} is {{NP-Hard}}.
\newblock {\em Journal of Machine Learning Research}, 2004.

\bibitem{weissteinAcyclicDigraph}
Eric~W. Weisstein.
\newblock Acyclic {{Digraph}}.
\newblock https://mathworld.wolfram.com/AcyclicDigraph.html, 2003.

\bibitem{jaynesPriorProbabilities1968}
Edwin~T. Jaynes.
\newblock Prior {{Probabilities}}.
\newblock {\em IEEE Transactions on Systems Science and Cybernetics}, 4(3):227--241, September 1968.

\bibitem{fristonPostHocBayesian2011}
Karl Friston and Will Penny.
\newblock Post hoc {{Bayesian}} model selection.
\newblock {\em NeuroImage}, 56(4):2089--2099, June 2011.

\bibitem{fristonBayesianModelReduction2019}
Karl Friston, Thomas Parr, and Peter Zeidman.
\newblock Bayesian model reduction.
\newblock {\em arXiv:1805.07092 [stat]}, October 2019.

\bibitem{ayInformationGeometry2017}
Nihat Ay, J{\"u}rgen Jost, H{\^o}ng~V{\^a}n L{\^e}, and Lorenz Schwachh{\"o}fer.
\newblock {\em Information {{Geometry}}}, volume~64 of {\em Ergebnisse Der {{Mathematik}} Und Ihrer {{Grenzgebiete}} 34}.
\newblock Springer International Publishing, Cham, 2017.

\bibitem{amariInformationGeometryIts2016}
S.~Amari.
\newblock {\em Information Geometry and Its Applications}.
\newblock Springer, 2016.

\bibitem{amariNaturalGradientWorks1998}
Shun-ichi Amari.
\newblock Natural {{Gradient Works Efficiently}} in {{Learning}}.
\newblock page~36, 1998.

\bibitem{winnVariationalMessagePassing2005}
John Winn and Christopher~M Bishop.
\newblock Variational {{Message Passing}}.
\newblock {\em Journal of Machine Learning Research}, page~34, 2005.

\bibitem{yedidiaConstructingFreeEnergyApproximations2005}
J.S. Yedidia, W.T. Freeman, and Y.~Weiss.
\newblock Constructing {{Free-Energy Approximations}} and {{Generalized Belief Propagation Algorithms}}.
\newblock {\em IEEE Transactions on Information Theory}, 51(7):2282--2312, July 2005.

\bibitem{murphyLoopyBeliefPropagation1999}
Kevin~P. Murphy, Yair Weiss, and Michael~I. Jordan.
\newblock Loopy belief propagation for approximate inference: An empirical study.
\newblock In {\em Proceedings of the {{Fifteenth}} Conference on {{Uncertainty}} in Artificial Intelligence}, {{UAI}}'99, pages 467--475, San Francisco, CA, USA, July 1999. Morgan Kaufmann Publishers Inc.

\bibitem{madiganBayesianGraphicalModels1995}
David Madigan, Jeremy York, and Denis Allard.
\newblock Bayesian {{Graphical Models}} for {{Discrete Data}}.
\newblock {\em International Statistical Review / Revue Internationale de Statistique}, 63(2):215--232, 1995.

\bibitem{giudiciImprovingMarkovChain2003}
Paolo Giudici and Robert Castelo.
\newblock Improving {{Markov Chain Monte Carlo Model Search}} for {{Data Mining}}.
\newblock {\em Machine Learning}, 50(1):127--158, January 2003.

\bibitem{eatonBayesianStructureLearning2012}
Daniel Eaton and Kevin Murphy.
\newblock Bayesian structure learning using dynamic programming and {{MCMC}}, June 2012.

\bibitem{grzegorczykImprovingStructureMCMC2008}
Marco Grzegorczyk and Dirk Husmeier.
\newblock Improving the structure {{MCMC}} sampler for {{Bayesian}} networks by introducing a new edge reversal move.
\newblock {\em Machine Learning}, 71(2):265--305, June 2008.

\bibitem{kuipersPartitionMCMCInference2017}
Jack Kuipers and Giusi Moffa.
\newblock Partition {{MCMC}} for {{Inference}} on {{Acyclic Digraphs}}.
\newblock {\em Journal of the American Statistical Association}, January 2017.

\bibitem{zhengDAGsNOTEARS2018}
Xun Zheng, Bryon Aragam, Pradeep~K Ravikumar, and Eric~P Xing.
\newblock {{DAGs}} with {{NO TEARS}}: {{Continuous Optimization}} for {{Structure Learning}}.
\newblock In {\em Advances in {{Neural Information Processing Systems}}}, volume~31. Curran Associates, Inc., 2018.

\bibitem{yuDAGGNNDAGStructure2019}
Yue Yu, Jie Chen, Tian Gao, and Mo~Yu.
\newblock {{DAG-GNN}}: {{DAG Structure Learning}} with {{Graph Neural Networks}}.
\newblock In {\em Proceedings of the 36th {{International Conference}} on {{Machine Learning}}}, pages 7154--7163. PMLR, May 2019.

\bibitem{lorchDiBSDifferentiableBayesian2021}
Lars Lorch, Jonas Rothfuss, Bernhard Sch{\"o}lkopf, and Andreas Krause.
\newblock {{DiBS}}: {{Differentiable Bayesian Structure Learning}}.
\newblock In {\em Advances in {{Neural Information Processing Systems}}}, volume~34, pages 24111--24123. Curran Associates, Inc., 2021.

\bibitem{liuSteinVariationalGradient2016}
Qiang Liu and Dilin Wang.
\newblock Stein {{Variational Gradient Descent}}: {{A General Purpose Bayesian Inference Algorithm}}.
\newblock In {\em Advances in {{Neural Information Processing Systems}}}, volume~29. Curran Associates, Inc., 2016.

\bibitem{saeediVariationalParticleApproximations2017}
Ardavan Saeedi, Tejas~D. Kulkarni, Vikash~K. Mansinghka, and Samuel~J. Gershman.
\newblock Variational {{Particle Approximations}}.
\newblock {\em Journal of Machine Learning Research}, 18(69):1--29, 2017.

\bibitem{mockusBayesianApproachGlobal1989}
Jonas Mockus.
\newblock {\em Bayesian {{Approach}} to {{Global Optimization}}: {{Theory}} and {{Applications}}}.
\newblock Springer Netherlands, Dordrecht, 1989.

\bibitem{lorchAmortizedInferenceCausal2022}
Lars Lorch, Scott Sussex, Jonas Rothfuss, Andreas Krause, and Bernhard Sch{\"o}lkopf.
\newblock Amortized {{Inference}} for {{Causal Structure Learning}}.
\newblock In {\em Advances in {{Neural Information Processing Systems}}}, October 2022.

\bibitem{deleuBayesianStructureLearning2022}
Tristan Deleu, Ant{\'o}nio G{\'o}is, Chris Emezue, Mansi Rankawat, Simon {Lacoste-Julien}, Stefan Bauer, and Yoshua Bengio.
\newblock Bayesian structure learning with generative flow networks.
\newblock In {\em Proceedings of the {{Thirty-Eighth Conference}} on {{Uncertainty}} in {{Artificial Intelligence}}}, pages 518--528. PMLR, August 2022.

\bibitem{deleuJointBayesianInference2023}
Tristan Deleu, Mizu {Nishikawa-Toomey}, Jithendaraa Subramanian, Nikolay Malkin, Laurent Charlin, and Yoshua Bengio.
\newblock Joint {{Bayesian Inference}} of {{Graphical Structure}} and {{Parameters}} with a {{Single Generative Flow Network}}.
\newblock {\em Advances in Neural Information Processing Systems}, 36:31204--31231, December 2023.

\bibitem{nishikawa-toomeyBayesianLearningCausal2022}
Mizu {Nishikawa-Toomey}, Tristan Deleu, Jithendaraa Subramanian, Yoshua Bengio, and Laurent Charlin.
\newblock Bayesian learning of {{Causal Structure}} and {{Mechanisms}} with {{GFlowNets}} and {{Variational Bayes}}, November 2022.

\bibitem{malkinGFlowNetsVariationalInference2022}
Nikolay Malkin, Salem Lahlou, Tristan Deleu, Xu~Ji, Edward~J. Hu, Katie~E. Everett, Dinghuai Zhang, and Yoshua Bengio.
\newblock {{GFlowNets}} and variational inference.
\newblock In {\em The {{Eleventh International Conference}} on {{Learning Representations}}}, September 2022.

\bibitem{cowanMagicalNumberShortterm2001}
N.~Cowan.
\newblock The magical number 4 in short-term memory: A reconsideration of mental storage capacity.
\newblock {\em The Behavioral and Brain Sciences}, 24(1):87--114; discussion 114--185, February 2001.

\bibitem{spelkeCoreKnowledge2007}
Elizabeth~S. Spelke and Katherine~D. Kinzler.
\newblock Core knowledge.
\newblock {\em Developmental Science}, 10(1):89--96, January 2007.

\bibitem{dacostaNeuralDynamicsActive2021}
Lancelot Da~Costa, Thomas Parr, Biswa Sengupta, and Karl Friston.
\newblock Neural {{Dynamics}} under {{Active Inference}}: {{Plausibility}} and {{Efficiency}} of {{Information Processing}}.
\newblock {\em Entropy}, 23(4):454, April 2021.

\bibitem{fristonActiveInferenceCuriosity2017}
Karl~J. Friston, Marco Lin, Christopher~D. Frith, Giovanni Pezzulo, J.~Allan Hobson, and Sasha Ondobaka.
\newblock Active {{Inference}}, {{Curiosity}} and {{Insight}}.
\newblock {\em Neural Computation}, 29(10):2633--2683, October 2017.

\bibitem{pavliotisMultiscaleMethodsAveraging2010}
Grigoris Pavliotis and Andrew Stuart.
\newblock {\em {Multiscale Methods: Averaging and Homogenization}}.
\newblock Springer, New York, NY, softcover reprint of hardcover 1st ed. 2008 {\'e}dition edition, November 2010.

\bibitem{tomovNeuralArchitectureTheorybased2023}
Momchil~S. Tomov, Pedro~A. Tsividis, Thomas Pouncy, Joshua~B. Tenenbaum, and Samuel~J. Gershman.
\newblock The neural architecture of theory-based reinforcement learning.
\newblock {\em Neuron}, 111(8):1331--1344.e8, April 2023.

\bibitem{wauthierSleepModelReduction2020}
Samuel~T Wauthier, Ozan {\c C}atal, Tim Verbelen, and Bart Dhoedt.
\newblock Sleep: {{Model Reduction}} in {{Deep Active Inference}}.
\newblock page~13, 2020.

\bibitem{smithActiveInferenceApproach2020}
Ryan Smith, Philipp Schwartenbeck, Thomas Parr, and Karl~J. Friston.
\newblock An {{Active Inference Approach}} to {{Modeling Structure Learning}}: {{Concept Learning}} as an {{Example Case}}.
\newblock {\em Frontiers in Computational Neuroscience}, 14, May 2020.

\bibitem{fristonSupervisedStructureLearning2023}
Karl~J. Friston, Lancelot Da~Costa, Alexander Tschantz, Alex Kiefer, Tommaso Salvatori, Victorita Neacsu, Magnus Koudahl, Conor Heins, Noor Sajid, Dimitrije Markovic, Thomas Parr, Tim Verbelen, and Christopher~L. Buckley.
\newblock Supervised structure learning, November 2023.

\bibitem{fristonPixelsPlanningScalefree2024}
Karl Friston, Conor Heins, Tim Verbelen, Lancelot Da~Costa, Tommaso Salvatori, Dimitrije Markovic, Alexander Tschantz, Magnus Koudahl, Christopher Buckley, and Thomas Parr.
\newblock From pixels to planning: Scale-free active inference, July 2024.

\bibitem{spelkeWhatBabiesKnow2022}
Elizabeth Spelke.
\newblock {\em What {{Babies Know}}: {{Core Knowledge}} and {{Composition Volume}} 1}.
\newblock Oxford University Press, New York, August 2022.

\bibitem{fristonVariationalSynthesisEvolutionary2023}
Karl Friston, Daniel~A. Friedman, Axel Constant, V.~Bleu Knight, Chris Fields, Thomas Parr, and John~O. Campbell.
\newblock A {{Variational Synthesis}} of {{Evolutionary}} and {{Developmental Dynamics}}.
\newblock {\em Entropy}, 25(7):964, July 2023.

\bibitem{lasotaChaosFractalsNoise1994}
Andrzej Lasota and Michael~C. MacKey.
\newblock {\em Chaos, {{Fractals}}, and {{Noise}}: {{Stochastic Aspects}} of {{Dynamics}}}.
\newblock Springer-Verlag, 1994.

\bibitem{dacostaUniversalInterpretableWorld2024}
Lancelot Da~Costa.
\newblock Toward {{Universal}} and {{Interpretable World Models}} for {{Open-ended Learning Agents}}, September 2024.

\bibitem{astromOptimalControlMarkov1965a}
K.~J {\AA}str{\"o}m.
\newblock Optimal control of {{Markov}} processes with incomplete state information.
\newblock {\em Journal of Mathematical Analysis and Applications}, 10(1):174--205, February 1965.

\bibitem{bartoReinforcementLearningIntroduction1992}
Andrew Barto and Richard Sutton.
\newblock {\em Reinforcement {{Learning}}: {{An Introduction}}}.
\newblock 1992.

\bibitem{brule2016computational}
Joshua Brul{\'e}.
\newblock The computational power of dynamic bayesian networks.
\newblock {\em arXiv preprint arXiv:1603.06125}, 2016.

\bibitem{parrNeuronalMessagePassing2019}
Thomas Parr, Dimitrije Markovic, Stefan~J. Kiebel, and Karl~J. Friston.
\newblock Neuronal message passing using {{Mean-field}}, {{Bethe}}, and {{Marginal}} approximations.
\newblock {\em Scientific Reports}, 9(1):1889, December 2019.

\bibitem{heinsCollectiveBehaviorSurprise2024}
Conor Heins, Beren Millidge, Lancelot Da~Costa, Richard~P. Mann, Karl~J. Friston, and Iain~D. Couzin.
\newblock Collective behavior from surprise minimization.
\newblock {\em Proceedings of the National Academy of Sciences}, 121(17):e2320239121, April 2024.

\bibitem{fristonHierarchicalModelsBrain2008}
Karl Friston.
\newblock Hierarchical {{Models}} in the {{Brain}}.
\newblock {\em PLoS Computational Biology}, 4(11):e1000211, November 2008.

\bibitem{hairerMarkovProcesses2020}
Martin Hairer and Xue-Mei Li.
\newblock Markov {{Processes}}.
\newblock January 2020.

\bibitem{fristonActiveInferenceIntentional2023a}
Karl~J. Friston, Tommaso Salvatori, Takuya Isomura, Alexander Tschantz, Alex Kiefer, Tim Verbelen, Magnus Koudahl, Aswin Paul, Thomas Parr, Adeel Razi, Brett Kagan, Christopher~L. Buckley, and Maxwell J.~D. Ramstead.
\newblock Active {{Inference}} and {{Intentional Behaviour}}, December 2023.

\bibitem{bakryAnalysisGeometryMarkov2014}
Dominique Bakry, Ivan Gentil, and Michel Ledoux.
\newblock {\em Analysis and {{Geometry}} of {{Markov Diffusion Operators}}}.
\newblock Grundlehren Der Mathematischen {{Wissenschaften}}. Springer International Publishing, 2014.

\bibitem{fristonWhatOptimalMotor2011}
Karl Friston.
\newblock What {{Is Optimal}} about {{Motor Control}}?
\newblock {\em Neuron}, 72(3):488--498, November 2011.

\bibitem{fristonActionUnderstandingActive2011}
Karl Friston, J{\'e}r{\'e}mie Mattout, and James Kilner.
\newblock Action understanding and active inference.
\newblock {\em Biological Cybernetics}, 104(1-2):137--160, February 2011.

\bibitem{lynnBrokenDetailedBalance2021}
Christopher~W. Lynn, Eli~J. Cornblath, Lia Papadopoulos, Maxwell~A. Bertolero, and Danielle~S. Bassett.
\newblock Broken detailed balance and entropy production in the human brain.
\newblock {\em arXiv:2005.02526 [cond-mat, physics:physics, q-bio]}, March 2021.

\bibitem{lindermanBayesianLearningInference2017}
Scott Linderman, Matthew Johnson, Andrew Miller, Ryan Adams, David Blei, and Liam Paninski.
\newblock Bayesian {{Learning}} and {{Inference}} in {{Recurrent Switching Linear Dynamical Systems}}.
\newblock In {\em Proceedings of the 20th {{International Conference}} on {{Artificial Intelligence}} and {{Statistics}}}, pages 914--922. PMLR, April 2017.

\bibitem{kidgerNeuralDifferentialEquations2022}
Patrick Kidger.
\newblock On {{Neural Differential Equations}}, February 2022.

\bibitem{parrDiscreteContinuousBrain2018}
Thomas Parr and Karl~J. Friston.
\newblock The {{Discrete}} and {{Continuous Brain}}: {{From Decisions}} to {{Movement}}---{{And Back Again}}.
\newblock {\em Neural Computation}, 30(9):2319--2347, September 2018.

\bibitem{bengioCurriculumLearning2009}
Yoshua Bengio, J{\'e}r{\^o}me Louradour, Ronan Collobert, and Jason Weston.
\newblock Curriculum learning.
\newblock In {\em Proceedings of the 26th {{Annual International Conference}} on {{Machine Learning}}}, {{ICML}} '09, pages 41--48, New York, NY, USA, June 2009. Association for Computing Machinery.

\bibitem{sovianyCurriculumLearningSurvey2022a}
Petru Soviany, Radu~Tudor Ionescu, Paolo Rota, and Nicu Sebe.
\newblock Curriculum {{Learning}}: {{A Survey}}.
\newblock {\em International Journal of Computer Vision}, 130(6):1526--1565, June 2022.

\bibitem{gershmanTutorialBayesianNonparametric2012}
Samuel~J. Gershman and David~M. Blei.
\newblock A tutorial on {{Bayesian}} nonparametric models.
\newblock {\em Journal of Mathematical Psychology}, 56(1):1--12, February 2012.

\bibitem{kahnemanThinkingFastSlow2013}
Daniel Kahneman.
\newblock {\em Thinking, {{Fast}} and {{Slow}}}.
\newblock {Farrar, Straus and Giroux}, New York, first edition edition, April 2013.

\bibitem{tscshantzHybridPredictiveCoding2023}
Alexander Tscshantz, Beren Millidge, Anil~K. Seth, and Christopher~L. Buckley.
\newblock Hybrid predictive coding: {{Inferring}}, fast and slow.
\newblock {\em PLOS Computational Biology}, 19(8):e1011280, August 2023.

\bibitem{bostrom2014superintelligence}
N.~Bostrom.
\newblock {\em Superintelligence: Paths, Dangers, Strategies}.
\newblock Oxford University Press, 2014.

\bibitem{russell2019human}
S.J. Russell.
\newblock {\em Human Compatible: Artificial Intelligence and the Problem of Control}.
\newblock Business book summary. Viking, 2019.

\bibitem{asimovRobot1950}
Isaac Asimov.
\newblock {\em I, {{Robot}}}.
\newblock Dennis Dobson, 1950.

\bibitem{kiversteinPlayfulnessMeaningfulLife2023}
Julian Kiverstein and Mark Miller.
\newblock Playfulness and~the meaningful life: An active inference perspective.
\newblock {\em Neuroscience of Consciousness}, 2023(1):niad024, January 2023.

\bibitem{waughResilienceAbilityMaintain2023}
Christian~E. Waugh and Anthony~W. Sali.
\newblock Resilience as the {{Ability}} to {{Maintain Well-Being}}: {{An Allostatic Active Inference Model}}.
\newblock {\em Journal of Intelligence}, 11(8):158, August 2023.

\bibitem{smithComputationalNeurosciencePerspective2022}
Ryan Smith, Lav~R. Varshney, Susumu Nagayama, Masahiro Kazama, Takuya Kitagawa, and Yoshiki Ishikawa.
\newblock A computational neuroscience perspective on subjective wellbeing within the active inference framework.
\newblock {\em International Journal of Wellbeing}, 12(4), October 2022.

\bibitem{millerPredictiveDynamicsHappiness2022}
Mark Miller, Julian Kiverstein, and Erik Rietveld.
\newblock The {{Predictive Dynamics}} of {{Happiness}} and {{Well-Being}}.
\newblock {\em Emotion Review}, 14(1):15--30, January 2022.

\bibitem{ueltzhofferVariationalFreeEnergy2021}
Kai Ueltzh{\"o}ffer, Lancelot Da~Costa, and Karl~J. Friston.
\newblock Variational free energy, individual fitness, and population dynamics under acute stress: {{Comment}} on ``{{Dynamic}} and thermodynamic models of adaptation'' by {{Alexander N}}. {{Gorban}} et al.
\newblock {\em Physics of Life Reviews}, 37:111--115, July 2021.

\bibitem{minkaDivergenceMeasuresMessage2005}
Thomas Minka.
\newblock Divergence measures and message passing.
\newblock Technical report, 2005.

\bibitem{frith_theory_2005}
Chris Frith and Uta Frith.
\newblock Theory of mind.
\newblock {\em Current Biology}, 15(17):R644--R645, September 2005.
\newblock Publisher: Elsevier.

\bibitem{fristonDuetOne2015}
Karl Friston and Christopher Frith.
\newblock A {{Duet}} for one.
\newblock {\em Consciousness and Cognition}, 36:390--405, November 2015.

\bibitem{ohataInvestigationSenseAgency2020}
Wataru Ohata and Jun Tani.
\newblock Investigation of the {{Sense}} of {{Agency}} in {{Social Cognition}}, based on frameworks of {{Predictive Coding}} and {{Active Inference}}: {{A}} simulation study on multimodal imitative interaction.
\newblock {\em arXiv:2002.01632 [cs]}, August 2020.

\bibitem{hyland2024freeenergy}
David Hyland, Tom{\'a}{\v{s}} Gaven{\v{c}}iak, Lancelot~Da Costa, Conor Heins, Vojtech Kovarik, Julian Gutierrez, Michael~J. Wooldridge, and Jan Kulveit.
\newblock Free-energy equilibria: Toward a theory of interactions between boundedly-rational agents.
\newblock In {\em ICML 2024 Workshop on Models of Human Feedback for AI Alignment}, 2024.

\bibitem{ashton2022problem}
Hal Ashton and Matija Franklin.
\newblock The problem of behaviour and preference manipulation in ai systems.
\newblock In {\em Ceur workshop proceedings}, volume 3087. CEUR Workshop Proceedings, 2022.

\bibitem{franklin2022recognising}
Matija Franklin, Hal Ashton, Rebecca Gorman, and Stuart Armstrong.
\newblock Recognising the importance of preference change: A call for a coordinated multidisciplinary research effort in the age of ai.
\newblock {\em arXiv preprint arXiv:2203.10525}, 2022.

\bibitem{Everitt2019UnderstandingAI}
Tom Everitt, Pedro~A. Ortega, Elizabeth Barnes, and Shane Legg.
\newblock Understanding agent incentives using causal influence diagrams. part i: Single action settings.
\newblock 2019.

\bibitem{everitt2021agentincentivescausalperspective}
Tom Everitt, Ryan Carey, Eric Langlois, Pedro~A Ortega, and Shane Legg.
\newblock Agent incentives: A causal perspective, 2021.

\bibitem{hylandFreeEnergyEquilibriaTheory2024}
David Hyland, Tom{\'a}{\v s} Gaven{\v c}iak, Lancelot Da~Costa, Conor Heins, Vojtech Kovarik, Julian Gutierrez, Michael~J. Wooldridge, and Jan Kulveit.
\newblock Free-{{Energy Equilibria}}: {{Toward}} a {{Theory}} of {{Interactions Between Boundedly-Rational Agents}}.
\newblock In {\em {{ICML}} 2024 {{Workshop}} on {{Models}} of {{Human Feedback}} for {{AI Alignment}}}, 2024.

\bibitem{isomuraBayesianFilteringMultiple2019}
Takuya Isomura, Thomas Parr, and Karl Friston.
\newblock Bayesian {{Filtering}} with {{Multiple Internal Models}}: {{Toward}} a {{Theory}} of {{Social Intelligence}}.
\newblock {\em Neural Computation}, 31(12):2390--2431, October 2019.

\bibitem{fristonFederatedInferenceBelief2024}
Karl~J. Friston, Thomas Parr, Conor Heins, Axel Constant, Daniel Friedman, Takuya Isomura, Chris Fields, Tim Verbelen, Maxwell Ramstead, John Clippinger, and Christopher~D. Frith.
\newblock Federated inference and belief sharing.
\newblock {\em Neuroscience \& Biobehavioral Reviews}, 156:105500, January 2024.

\bibitem{ward2024reasons}
Francis~Rhys Ward, Matt MacDermott, Francesco Belardinelli, Francesca Toni, and Tom Everitt.
\newblock The reasons that agents act: Intention and instrumental goals.
\newblock In {\em Proceedings of the 23rd International Conference on Autonomous Agents and Multiagent Systems}, AAMAS '24, page 1901–1909, Richland, SC, 2024. International Foundation for Autonomous Agents and Multiagent Systems.

\bibitem{ward2024honesty}
Francis Ward, Francesca Toni, Francesco Belardinelli, and Tom Everitt.
\newblock Honesty is the best policy: defining and mitigating ai deception.
\newblock {\em Advances in Neural Information Processing Systems}, 36, 2024.

\bibitem{Krakovna2018PenalizingSE}
Victoria Krakovna, Laurent Orseau, Miljan Martic, and Shane Legg.
\newblock Penalizing side effects using stepwise relative reachability.
\newblock {\em arXiv: Learning}, 2018.

\bibitem{baker_bayesian_2011}
Chris~L. Baker, R.~Saxe, and J.~Tenenbaum.
\newblock Bayesian {Theory} of {Mind}: {Modeling} {Joint} {Belief}-{Desire} {Attribution}.
\newblock {\em Cognitive Science}, 2011.

\bibitem{kleiman-weiner_evolving_2025}
Max Kleiman-Weiner, Alejandro Vientós, David~G. Rand, and Joshua~B. Tenenbaum.
\newblock Evolving general cooperation with a {Bayesian} theory of mind.
\newblock {\em Proceedings of the National Academy of Sciences}, 122(25):e2400993122, June 2025.
\newblock Publisher: Proceedings of the National Academy of Sciences.

\bibitem{ramsteadBayesianMechanicsPhysics2022}
Maxwell J.~D. Ramstead, Dalton A.~R. Sakthivadivel, Conor Heins, Magnus Koudahl, Beren Millidge, Lancelot Da~Costa, Brennan Klein, and Karl~J. Friston.
\newblock On {{Bayesian Mechanics}}: {{A Physics}} of and by {{Beliefs}}, May 2022.

\bibitem{dacostaBayesianMechanicsMetacognitive2024}
Lancelot Da~Costa and Lars {Sandved-Smith}.
\newblock Towards a {{Bayesian}} mechanics of metacognitive particles: {{A}} commentary on ``{{Path}} integrals, particular kinds, and strange things'' by {{Friston}}, {{Da Costa}}, {{Sakthivadivel}}, {{Heins}}, {{Pavliotis}}, {{Ramstead}}, and {{Parr}}.
\newblock {\em Physics of Life Reviews}, 48:11--13, March 2024.

\bibitem{sandved-smithMetacognitiveParticlesMental2024}
Lars {Sandved-Smith} and Lancelot Da~Costa.
\newblock Metacognitive particles, mental action and the sense of agency, May 2024.

\bibitem{hafnerActionPerceptionDivergence2020}
Danijar Hafner, Pedro~A. Ortega, Jimmy Ba, Thomas Parr, Karl Friston, and Nicolas Heess.
\newblock Action and {{Perception}} as {{Divergence Minimization}}.
\newblock {\em arXiv:2009.01791 [cs, math, stat]}, October 2020.

\bibitem{lecunPathAutonomousMachine2022}
Yann LeCun.
\newblock A {{Path Towards Autonomous Machine Intelligence Version}} 0.9.2, 2022-06-27.
\newblock June 2022.

\bibitem{bengioFlowNetworkBased2021}
Emmanuel Bengio, Moksh Jain, Maksym Korablyov, Doina Precup, and Yoshua Bengio.
\newblock Flow {{Network}} based {{Generative Models}} for {{Non-Iterative Diverse Candidate Generation}}.
\newblock In {\em Advances in {{Neural Information Processing Systems}}}, volume~34, pages 27381--27394. Curran Associates, Inc., 2021.

\end{thebibliography}

\end{document}